\documentclass[conference]{IEEEtran}
\usepackage{times}
\usepackage{amsmath,amsfonts,bm}
\usepackage{todonotes}
\newcommand{\ttt}[1]{\texttt{\small #1}}

\def\eqref#1{equation~\ref{#1}}
\def\1{\bm{1}}

\def\vu{{\bm{u}}}
\def\vv{{\bm{v}}}

\def\mC{{\bm{C}}}

\def\mT{{\bm{T}}}

\def\mW{{\bm{W}}}

\DeclareMathAlphabet{\mathsfit}{\encodingdefault}{\sfdefault}{m}{sl}
\SetMathAlphabet{\mathsfit}{bold}{\encodingdefault}{\sfdefault}{bx}{n}

\def\gS{{\mathcal{S}}}

\def\sR{{\mathbb{R}}}

\usepackage{algorithm}
\usepackage{algorithmic}
\usepackage[hidelinks,colorlinks=true,linkcolor=blue,citecolor=blue,urlcolor=blue,filecolor=blue]{hyperref}
\usepackage{url}
\usepackage{wrapfig}
\usepackage{subcaption}
\let\labelindent\relax
\usepackage{enumitem}
\usepackage{color}
\definecolor{RoyalBlue}{rgb}{0.12, 0.50, 0.69} % Approximate Royal Blue

\usepackage{booktabs}
\usepackage{array}
\usepackage[numbers]{natbib}
\usepackage{multicol}

\begin{document}

% paper title
\newcommand{\OUR}{\textsc{Sketch-To-Skill}}

\title{Sketch-to-Skill: Bootstrapping Robot Learning with Human Drawn Trajectory Sketches}
% \title{Template paper for the \\Robotics: Science and Systems Conference}

% You will get a Paper-ID when submitting a pdf file to the conference system
%\author{Author Names Omitted for Anonymous Review. Paper-ID [98]}

%\author{\authorblockN{Michael Shell}
%\authorblockA{School of Electrical and\\Computer Engineering\\
%Georgia Institute of Technology\\
%Atlanta, Georgia 30332--0250\\
%Email: mshell@ece.gatech.edu}
%\and
%\authorblockN{Homer Simpson}
%\authorblockA{Twentieth Century Fox\\
%Springfield, USA\\
%Email: homer@thesimpsons.com}
%\and
%\authorblockN{James Kirk\\ and Montgomery Scott}
%\authorblockA{Starfleet Academy\\
%San Francisco, California 96678-2391\\
%Telephone: (800) 555--1212\\
%Fax: (888) 555--1212}}

% avoiding spaces at the end of the author lines is not a problem with
% conference papers because we don't use \thanks or \IEEEmembership

% for over three affiliations, or if they all won't fit within the width
% of the page, use this alternative format:
% 
\author{\authorblockN{Peihong Yu\authorrefmark{1}\authorrefmark{2},
Amisha Bhaskar\authorrefmark{1}\authorrefmark{2},
Anukriti Singh\authorrefmark{1}, 
Zahiruddin Mahammad\authorrefmark{1} and
Pratap Tokekar\authorrefmark{1}}
\authorblockA{\authorrefmark{1}Department of Computer Science\\
University of Maryland,
College Park, Maryland, 20742\\ 
Emails: \{peihong, amishab, anukriti, zahirmd, tokekar\}@umd.edu\\}
\authorblockA{\authorrefmark{2}Peihong Yu and Amisha Bhaskar contributed equally to this work. }
}

\maketitle

\begin{abstract}

Training robotic manipulation policies traditionally requires numerous demonstrations and/or environmental rollouts. While recent Imitation Learning (IL) and Reinforcement Learning (RL) methods have reduced the number of required demonstrations, they still rely on expert knowledge to collect high-quality data, limiting scalability and accessibility. We propose \OUR{}, a novel framework that leverages human-drawn 2D sketch trajectories to bootstrap and guide RL for robotic manipulation. Our approach extends beyond previous sketch-based methods, which were primarily focused on imitation learning or policy conditioning, limited to specific trained tasks. \OUR{} employs a Sketch-to-3D Trajectory Generator that translates 2D sketches into 3D trajectories, which are then used to autonomously collect initial demonstrations. We utilize these sketch-generated demonstrations in two ways: to pre-train an initial policy through behavior cloning and to refine this policy through RL with guided exploration. Experimental results demonstrate that \OUR{} achieves 
% comparable performance to traditional approaches that leverage high-quality demonstration data, while using only simple 2D sketches as input. \OUR{}  achieves 
$\sim$96\% of the performance of the baseline model that leverages teleoperated demonstration data, while exceeding the performance of a pure reinforcement learning policy by $\sim$170\%, only from sketch inputs. This makes robotic manipulation learning more accessible and potentially broadens its applications across various domains.
% \footnote{Anonymized code and demo datasets are available on \href{https://gsalerts-cyber.github.io/sketch-to-skill/}{our webpage}.}

% Our work addresses a crucial gap by integrating sketches into RL, extending their application beyond imitation learning and policy conditioning.

% We propose a novel approach using human-drawn 2D sketch trajectories as demonstrations, significantly lowering the entry barrier for robotic learning. Our method employs a Sketch-to-3D Trajectory Generator that learns to translate 2D sketches into 3D trajectories,  which can be used as surrogate for expert demonstration data., based on which we then initialize a robotic policy through behavior cloning. We then train the policy with RL, incorporating a discriminator for guided exploration that encourages the policy to generate trajectories closer to those produced by the Sketch-to-Trajectory Generator. Experimental results show comparable performance to traditional methods that leverage high-quality demonstration data, while our approach uses only simple 2D sketches. This makes robotic manipulation learning more accessible and potentially broadens its applications across various domains.

% Replace last lines with real numbers whenever you have them

\end{abstract}

% 3-5 is *very* *very* task specific
% It's only for the simple tasks. Complex tasks still need 100s of demos. So I suggest we don't mention a specific number here.
% Instead, I would say something like..
% A limited number of expert demonstrations can be used to bootstrap reinforcement learning, which brings down the number of total environmental rollouts. However, gathering expert demonstrations still relies on expert knowle..

% I suggest mentioning RT-sketch here.
% Basically something like RT-Sketch showed how 2D trajectory sketches improves the generalization of IL. However, that was limited only to IL. We show how to use sketches in IL

% We should clarify that there are two ways we use sketch generated trajectories. As imitation data to bootstrap RL and for guiding exploration during RL.

% We should differentiate between robot expert and domain expert. 
% Maybe hard to do in the abstract but certainly in the intro.
% Traditional: robot experts teleop or control the robot to provide robot demo data.
% Sketches: domain expert but not necessarily robot expert provide sketches. non-robot expert because they don't need to know how to control the robot just what the trajectory should look like.
% See "Demonstrator Expertise" para in https://arxiv.org/pdf/2311.16098
% Not easy for novice data collectors to collect high quality data right at the start. They need some training to use the system

\IEEEpeerreviewmaketitle

\section{Introduction}

% problem definition

\begin{figure*}
    \centering
    \includegraphics[width=0.95\linewidth]{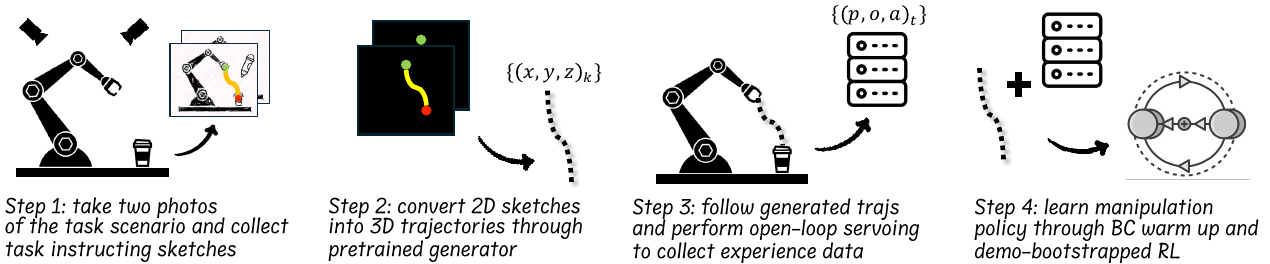}
    \caption{Learning a new skill in the \OUR{} framework. Step 1: Capture the task scenario from two views and collect human-drawn sketches. Step 2: Convert 2D sketches to 3D trajectories using a pretrained generator. Step 3: Execute generated trajectories to collect experience data. Step 4: Learn manipulation policy using reinforcement learning bootstrapping from  behavior cloning and using  guidance for experience data.}
    \label{fig:overview}
\end{figure*}

% figures in the shared slides
Robots are increasingly being deployed in dynamic environments, where they must perform a wide range of tasks with precision and adaptability. One of the key challenges in enabling robots to learn new skills lies in specifying complex, task-specific behaviors. Learning from Demonstration (LfD) (\cite{Billard:2013}) has become a widely used approach, allowing robots to acquire novel motions by imitating expert-provided trajectories.
However, collecting demonstration data for LfD is challenging, particularly for high degree-of-freedom (DOF) robots performing manipulation. 

% Traditional methods such as kinesthetic teaching and teleoperation while useful also have challenges with safety risks, scalability, and the need for specialized expertise \citep{chan2014application, ferraguti2015bilateral, bimbo2017teleoperation}. Recent approaches, such as using manually-operated grippers instrumented with smartphone apps~\citep{shafiullah2023bringing} and Virtual Reality (VR) based teleoperation systems~\citep{kamijo2024learning}, offer more intuitive hardware interfaces for collecting demonstrations. However, they require specialized hardware which may limit their flexibility and accessibility. Recently there has been interest in leveraging an innate human ability to communicate spatial ideas and motions through simple sketches. For example, a quick sketch of a path can easily communicate the intended movement for navigating toward a goal location. 

While methods like kinesthetic teaching and teleoperation \citep{chan2014application, ferraguti2015bilateral, bimbo2017teleoperation} are well-established and effective approaches for collecting demonstrations, the robotics community continues to explore complementary methods to expand the toolkit available for robot learning. Several new approaches have emerged, such as using manually-operated grippers instrumented with smartphone apps~\citep{shafiullah2023bringing} and Virtual Reality (VR) based teleoperation systems~\citep{kamijo2024learning}, offering more intuitive hardware interfaces for collecting demonstrations. There has also been growing interest in leveraging an innate human ability to communicate spatial ideas and motions through simple sketches. For example, a quick sketch of a path can easily communicate the intended movement for navigating toward a goal location.

Researchers have begun to explore this promising direction. RT-Trajectory \citep{gu2023rt} introduced the notion of sketches and showed how to use coarse trajectory sketches for policy conditioning in Imitation Learning (IL). RT-Sketch \citep{sundaresan2024rt} extended this concept to leverage hand-drawn sketches of the entire environment for goal-conditioned IL. These methods demonstrated the potential of utilizing sketches in robotics, but they were primarily focused on IL and biased towards tasks they were specifically trained on. \citet{zhi2023learning} expanded this idea with \textit{diagrammatic teaching}, where users instruct robots by sketching motion trajectories directly on 2D images of the scene. Their approach uses density estimation and ray tracing to reconstruct 3D trajectories from the sketches, thus limiting its ability to replicate only the provided sketches and restricting its generalization to new or unseen task setups.

Unlike prior work that used sketches only as conditioning in IL, we present a more {generalizable} approach that learns to predict 3D trajectories from sketches in Reinforcement Learning (RL).
Specifically, we propose \OUR{} (Figure \ref{fig:overview}), a framework that  bootstraps and guides RL using sketches. Our approach first learns to map 2D sketches to 3D trajectories, which are then used to collect demonstrations. We utilize these sketch-generated demonstrations in two ways: first, by pre-training an initial policy through Behavior Cloning (BC), and second, by refining this policy through reinforcement learning with guided exploration. Although sketch-generated demonstrations are not as precise or high-quality as teleoperated ones, they still contain enough useful information to aid RL and reduce learning time.

% Sketch-generated demonstrations offer valuable guidance but are not as precise or high-quality as teleoperated demonstrations. They also suffer from partial observability, as they fail to capture detailed gripper motions or fine-grained manipulation strategies. As a result, they cannot be used directly in imitation learning (IL). However, despite these limitations, sketches still contain enough useful information to aid reinforcement learning (RL) and reduce learning time. We leverage them in two key ways:

% Unlike teleoperation which requires specialized hardware and proficiency in using the system, sketches can generally be provided by non-robotics experts. 
% This approach capitalizes on the accessibility of sketching, allowing contributions from non-experts and significantly reducing the barriers typically associated with teleoperation systems.  By treating these sketch-based trajectories as approximate guiding signals rather than high-fidelity demonstrations, we allow the agent to learn more effectively even with coarse sketches. 
This approach capitalizes on the accessibility of sketching, allowing contributions from non-experts without requiring specialized hardware. By treating these sketch-based trajectories as approximate guiding signals, we demonstrate that even simple 2D sketches can enable policies that achieve performance comparable to those trained with teleoperated demonstrations.
We summarize our contributions as follows:

% This is where our approach, \textit{Sketch-to-Skill}, comes in. We propose a novel framework that uses human-drawn 2D sketches as a starting point for reinforcement learning (RL). While sketch-generated demonstrations are inherently lower in quality than real-world demos, they still contain valuable signals that can guide and bootstrap RL, thereby reducing the learning time required. Although sketches may lack fine details such as gripper motions or manipulation strategies, they effectively provide useful guidance for RL. We leverage these sketches in two key ways:

% \textbf{1. Bootstrapping RL:} Sketches provide a rough outline of task objectives, helping to initialize the agent’s learning process and guide its early exploration.

% \textbf{2. Guided Exploration:} Throughout the RL process, sketch-generated trajectories direct the agent's exploration, focusing it on relevant areas and improving learning efficiency, even though the sketches are not treated as exact ground truth.

% By using these sketch-based trajectories as guiding signals rather than ground truth, we enable the agent to learn more effectively. Our contributions are as follows:

\begin{enumerate}[label=(\arabic*), leftmargin=1cm, labelsep=0.2cm, itemsep=0pt] 
\item We identify and address a crucial gap by integrating sketches into RL, extending their application beyond imitation learning and policy conditioning. 
\item We propose \OUR, a framework that leverages sketches to bootstrap and guide RL, reducing reliance on high-quality, real-world demonstrations. 
\item Through extensive experiments, we demonstrate that sketches, despite their low fidelity, significantly accelerate learning by improving exploration and task comprehension in RL. \OUR{}  achieves $\sim$96\% of the performance of the baseline model utilizing high-quality teleoperation demonstrations, while exceeding the performance of a pure reinforcement learning policy by $\sim$170\% during evaluation. 
% \PRT{add punchline numbers here!}
\end{enumerate}

% To address these challenges, we propose an alternative method using human-drawn 2D sketches as demonstrations. By translating sketches into robot trajectories, we significantly reduce the reliance on traditional demonstration methods while maintaining the quality necessary for robotic learning. This approach offers a scalable solution for generating diverse demonstration data, particularly for tasks involving complex, multi-step trajectories.

% With this trained model, we can significantly lower the barrier to entry for demonstrating robotic manipulation tasks. It enables non-experts to intuitively guide robot learning through simple 2D sketches, effectively bridging the gap between human intent and robotic execution. In the context of robot task learning, human-drawn sketches can be used as input to obtain 3D trajectories, making the process more accessible and user-friendly. Our approach generalizes to various sketching styles, making it versatile for practical applications.

% For a practical example, consider a pick-and-place task in a warehouse setting. Instead of programming precise coordinates or physically guiding the robot, a user could simply sketch the desired trajectory on two images of the workspace. Our system would then generate a smooth, executable 3D path for the robot, significantly streamlining the task specification process.

% \PRT{Anukriti/All: LfD, Imitation bootstrapper RL, Papers that provide alternative methods to collect demos... VR, exoskeleton suits, Sticks, learning from human videos etc. Sketch papers}
\section{Related Work}

% \subsection{Learning from Demonstration}
\textbf{Learning from Demonstration (LfD).}
LfD~\citep{Billard:2013} is a key method in robot learning, allowing robots to acquire skills through expert demonstrations, bypassing the complexities of action programming and cost function design \citep{Ravichandar2020RecentAI}. Kinesthetic teaching, where an expert physically guides the robot while its movements are recorded, is widely used in methods like DMPs \citep{Kober_2009, Ijspeert_2013}, Probabilistic Movement Primitives \citep{Paraschos_2015}, and stable dynamical systems \citep{Khansari_Zadeh_2011, Mohammad_Khansari_Zadeh_2014, Bevanda_2022}. However, it is labor-intensive and challenging to scale. 
% LfD~\citep{Billard:2013} is a widely adopted technique in robot learning, enabling robots to acquire new skills by observing expert demonstrations. This approach circumvents the need for detailed action programming or manual cost function design, both of which can be complex and time-consuming \citep{Ravichandar2020RecentAI}. One of the most common methods for obtaining demonstrations is kinesthetic teaching, where an expert physically moves the robot through the desired motion while its joint or end-effector positions are recorded. Approaches such as Dynamical Movement Primitives (DMPs) \citep{Kober_2009, Ijspeert_2013}, Probabilistic Movement Primitives \citep{Paraschos_2015}, and stable dynamical system models \citep{Khansari_Zadeh_2011, Mohammad_Khansari_Zadeh_2014, Bevanda_2022} have utilized this technique. Despite its effectiveness, kinesthetic teaching requires the user to physically interact with the robot, making it labor-intensive and challenging to scale.
Teleoperation~\citep{si2021review}, where users control robots remotely, offers more flexibility but can be complex and requires expertise to operate. VR interfaces~\citep{zhang2018deep,kamijo2024learning} provide a more immersive alternative but depend on specialized hardware. To overcome these limitations, recent research has introduced more accessible approaches, like sketch-based demonstrations \citep{drolet2024comparison}.
% An alternative approach to collecting demonstrations is teleoperation~\citep{si2021review}, where the user controls the robot remotely, often with a handheld device or controller. Teleoperation allows for more flexibility, as the user can provide demonstrations without being physically near the robot \citep{wu2023gello}. However, this method is not without its difficulties, as remote control can be complex and requires proficiency. To alleviate some of these challenges, VR interfaces~\citep{zhang2018deep,kamijo2024learning} have been introduced, allowing users to interact with the robot in a more immersive environment, though these systems require specialized hardware. 

% While traditional methods like kinesthetic teaching and teleoperation have proven useful in LfD, they come with significant limitations in terms of accessibility and scalability \citep{drolet2024comparison}. To address these challenges, recent work has focused on more accessible interfaces to provide demonstrations, such as sketches.

% approaches such as Diagrammatic Teaching~\citep{zhi2023learning} offer a more intuitive alternative by allowing users to sketch demonstrations directly over 2D images. This approach provides a less hardware-dependent and more user-friendly solution for generating robot demonstrations.

% \subsection{Sketches in Robotics}
\textbf{Sketches in Robotics.}
Sketches have become a powerful tool in computer vision, aiding tasks like scene understanding \citep{chowdhury2023scenetrilogy} and object detection \citep{chowdhury2023can, bhunia2023sketch2saliency}. RT-Sketch \citep{sundaresan2024rt} first explored hand-drawn sketches for goal-conditioned imitation learning (IL), using them to define tasks intuitively. RT-Trajectory \citep{gu2023rt} extended this by using trajectory sketches as IL policy conditioning, either drawn by users or generated by a Large Language Model from task descriptions. Similarly, the Diagrammatic Teaching framework \citep{zhi2023learning} uses density estimation and ray tracing to reconstruct 3D trajectories from the sketches. These methods, however, only use sketches as conditioning for task completion, and thus do not generalize beyond the tasks where the sketches are provided.
\textbf{Demonstration-Enhanced Strategies for Efficient RL.} 
Incorporating demonstration data in RL can improve sample efficiency, especially in environments where rewards are sparse. Methods such as Reinforcement Learning from Prior Data (RLPD) \citep{smith2022walk}, Imitation Bootstrapped RL (IBRL) \citep{hu2023imitation} and PLANRL \citep{bhaskar2024planrl} take advantage of prior demonstrations by embedding them into the agent’s replay buffer. During training, these examples are oversampled, offering the agent more frequent exposure to expert-guided trajectories. Such approaches significantly improve learning speed and performance, particularly in continuous control tasks where learning from scratch can be prohibitively slow and inefficient \citep{yu2024beyond}.
Our research expands upon these  techniques by exploring how sketch-based trajectories can be used as an additional source of prior data in RL. %By integrating sketches into the RL framework, we aim to enhance the agent’s adaptability in new environments. This combination of sketch-guided learning with existing RL techniques has the potential to provide a more efficient path to mastering complex manipulation tasks. \yuph{this para to be revised}

\section{\OUR{}}

% \PRT{Peihong, I think we need a problem statement here. Something that describes what the inputs are (using the notation in the pseudocode) and what seek as the output. You can use the one in Section 3 of https://openreview.net/pdf?id=YxvmODVWny as reference}

Our approach bootstraps robot learning from trajectory sketches, significantly lowering the barrier to entry for robotic task specification. This section details our three-stage method: (1) training a Sketch-to-3D Trajectory Generator, (2) obtaining 3D trajectories and execution experiences through the Generator and open-loop servoing,  (3) pre-training an initial robotic manipulation policy through behavior cloning, and refining the policy through reinforcement learning with guided exploration. By integrating intuitive human input with powerful learning algorithms, our approach aims to create more accessible and adaptable robotic learning systems.

% \PRT{Also, I think we need some notation. Specifically, we use terms sketch generated trajectory, collected trajectory, open loop trajectory and that can be confusing. Maybe would help to have some notation for each one. I think we'll need something for
% - sketch inputs
% - latent in the teacher model
% - trajectory generated from the teacher outputs
% - open loop servo trajectory
% - collected demonstrations using the open loop servo trajectory (which is not just the trajectory but the actual demo)}

\subsection{Sketch-to-3D Trajectory Generator}

% <highlight>Our Sketch-to-3D Trajectory Generator represents a novel approach to robotic trajectory generation, bridging the gap between intuitive human input and precise robotic execution. This section outlines the key components of our method: data generation, network architecture, trajectory representation, and training process. We also discuss the advantages and potential limitations of our approach.</highlight>

\begin{figure*}
    \centering
    \includegraphics[width=0.65\linewidth]{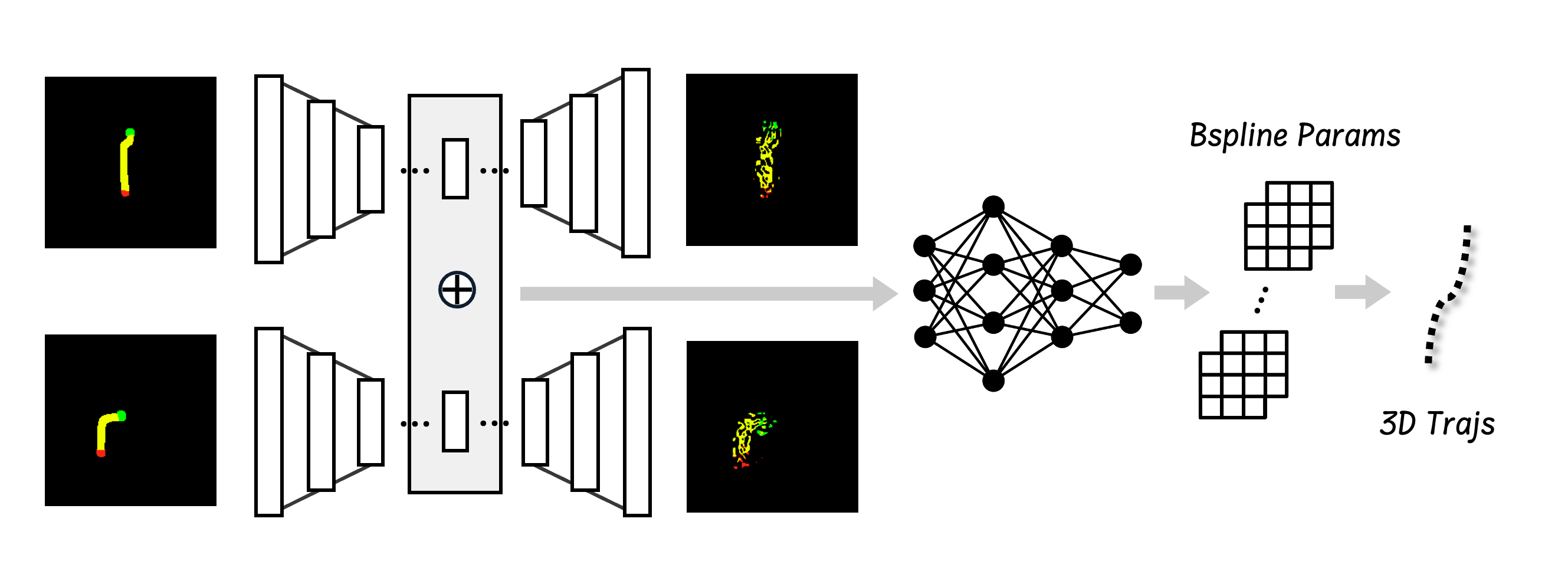}
    \caption{The Sketch-to-3D Trajectory Generator takes dual-view 2D sketches as inputs and predicts B-spline parameters to generate the final 3D trajectory output. }
    \label{fig:teacher}
\end{figure*}

Our method begins with a Sketch-to-3D Trajectory Generator, $\mT$, that translates a pair of 2D sketch images $(I^{1}, I^{2})$ obtained from different viewpoints into corresponding 3D robot trajectory $\xi_g$. To train this generator, we use a dataset  consisting of 3D robot end-effector trajectories along with their 2D sketches from two viewpoints. 
% These trajectories can be obtained from various sources, such as play data where the robot executes sequences of actions. 
These trajectories can be obtained from various sources, such as robot arm play data in simulation or real world where the robot executes sequences of actions, or existing recorded trajectory datasets transformed to match the workspace coordinates. 
Sketches during inference can be provided by a human on RGB images of the scene, as shown in Figure \ref{fig:hand-drawn}. However, the sketches fed as input to the generator are 2D projections on blank backgrounds, with green and red dots representing the start and end points respectively, and yellow lines for the trajectory (see Figure \ref{fig:teacher} for an example). By focusing solely on the trajectory information without additional scene complexity, our model can efficiently learn to encode the dual-view sketches and decode them into the corresponding 3D trajectory.

% To train this generator, we create a large dataset using simulation or utilize existing trajectory datasets when available. For simulated data, we generate random 3D trajectories in a simulated environment and project them onto predefined camera views, producing training pairs of 3D trajectories and their 2D projections from two viewpoints. Note that the 2D projections are pure sketches on blank backgrounds, with green and red dots representing the start and end points respectively, and yellow lines for the trajectory.

% \PRT{I don't think you should differentiate simulation and real world trajs here. Just say what dataset we need to train the Sketch-To-Traj model. To train this generator, we use a dataset consisting of 3D robot end-effector trajectories along with the 2D sketches from two viewpoints. Note that the 2D projections are ... (refer to an example figure here) These trajectories can be obtained from, for example, \emph{play} data, where the robot executes open-loop sequence of actions.}

The generator uses a neural network to map dual-view 2D sketches to 3D trajectories, where we adopt a hybrid architecture combining a Variational Autoencoder (VAE) \citep{kingma2013auto} and a Multilayer Perceptron (MLP), as illustrated in Figure \ref{fig:teacher}. The VAE encodes sketches from two  viewpoints, ideally orthogonal, to resolve depth ambiguity and capture essential trajectory features. 
% This dual-view approach ensures the model efficiently learns to translate sketches into accurate 3D trajectories, focusing solely on the trajectory data without added scene complexity.
% The generator employs a neural network-based architecture designed to learn the mapping from dual-view 2D sketches to their corresponding 3D trajectories. Given the simplified nature of our input data, we adopt a streamlined hybrid architecture combining a Variational Autoencoder (VAE) with a Multilayer Perceptron (MLP), as visualized in Figure \ref{fig:teacher}. The VAE encoder processes sketches from two viewpoints, creating a latent space that captures the essential features of the sketched trajectories. These two views are set up to provide clear, distinct perspectives of the trajectory, ideally from orthogonal angles to maximize information capture and minimize occlusions. This dual-view approach is key to resolving the depth ambiguity inherent in single-view 2D sketches. By focusing solely on the trajectory information without additional scene complexity, our model can efficiently learn to encode the dual-view sketches and decode them into the corresponding 3D trajectory.
% \PRT{% uses a Variational Autoencoder (VAE) that maps the two 2D sketches to their..refer to figure 2, somewhere you should mention what the two views should be.. ideally we would have some numbers telling the effect of varous combination of views}
% \PRT{Might need to condense some of this and move to appendix in the interest of space}
The MLP decoder generates B-spline~\citep{prautzsch2002bezier} control points $\mC \in \sR^{n_{cp} \times 3}$ from the latent representation, which we then use to interpolate smooth 3D trajectories. 
% B-splines, defined by a knot vector, control points, and the spline degree, offer the flexibility to generate trajectories of varying density from the same parameters, making our method adaptable to different task requirements.
% The B-splines are defined by a knot vector, control points, and the spline degree, determining the trajectory's shape and continuity.
% We chose B-splines for their ability to express complex paths with relatively few parameters while allowing for sequential trajectory generation with various resolutions.
% The flexibility of B-splines allows us to generate trajectories of varying density from the same parameters, making our method adaptable to different task requirements.
% \PRT{Need a sentence explaining general how B-Splines represent a trajectory..Like B-splines implicitly represent a 3D trajectory in the form of control points and xx parameters. The MLP trained on the encoded sketches is trained to predict these parameters.To reduce the ..}
We adopted uniform knots and pre-compute the B-spline parametrization matrix $\mW \in \sR^{n_{tp} \times n_{cp}}$ to reduce computational complexity and facilitate efficient backpropagation. 
% To reduce computational complexity and facilitate efficient backpropagation, we adopt uniform knots in our B-spline formulation. The MLP, trained on the encoded sketches, learns to predict only the control points. This choice allows us to pre-compute two key components for efficient trajectory generation: (1) the uniform knot vector $\vu$, which depends on the degree of the B-spline and the number of control points $n_{cp}$, (2) 
The calculation of $\mW$ only depends on the uniform knot vector $\vu$ and the desired number of points $n_{tp}$ in the generated trajectory, and can be pre-calculated using the Cox-de Boor recursion formula (also known as de Boor's algorithm \cite{deboor-doi:10.1137/0714026}, see Appendix \ref{appen:deboor} for details). Then the final trajectory generation is simply a matrix multiplication: $\xi_g \in \sR^{n_{tp} \times 3} = \mW \cdot \mC$. With the generated control point parameter $\mC$, we can also easily generate trajectories of varying density from the same parameters, making our method adaptable to different task requirements.

% Our training process uses a multi-component loss function $L=L_{traj}+L_{sketch}+L_{kld}$, where $L_{traj}$ represents the trajectory reconstruction error, $L_{sketch}$ represents the reconstruction error of both sketch images, and $L_{kld}$ is the KL-divergence regularization term applied to the latent feature vectors (as shown in Figure \ref{fig:teacher}). This ensures accurate trajectory generation while maintaining sketch fidelity and a well-structured latent space. To enhance robustness and generalization, our training process employs two concurrent data augmentation strategies. The first applies diverse image augmentations (rotations, scaling, affine transformations, noise) to input sketches, used exclusively for updating the VAE to learn robust sketch representations. The second strategy targets potential mismatches in hand-drawn sketches by subtly modifying both original sketches and their 3D trajectories. This involves adding noise and minor elastic deformations to sketches, and noise with refitting to trajectories. These augmented pairs update the entire model, preparing it for hand-drawn input variability while maintaining sketch-trajectory consistency. This augmentation approach enhances the model's ability to handle diverse, imperfect sketches while ensuring accurate 3D trajectory generation in real-world scenarios. 
Our training process uses a multi-component loss function $L=L_{traj}+L_{sketch}+L_{kld}$, where $L_{traj}$ handles trajectory reconstruction, $L_{sketch}$ manages sketch reconstruction (Mean Square Error), and $L_{kld}$ is KL-divergence for latent space regularization (Figure \ref{fig:teacher}). This ensures accurate trajectory generation while preserving sketch fidelity and latent space structure. We also applied data augmentation to both the sketch images and the trajectories to enhance the model's robustness and generalization (more details can be found in Appendix \ref{appen:traj_app}).

We can use the trained Sketch-to-3D Trajectory Generator, $\mT$, to generate demonstrations $\{\xi_D\}$ for learning new tasks using sketches drawn by a human. Specifically, the human draws trajectory sketches on two views of RGB images captured from the initial task state. This is similar to how human-drawn sketches are generated in prior works~\citep{gu2023rt, zhi2023learning}. These paired sketch images, $\{(I^{1}, I^{2})\}$, are input into our trained generator, which produces corresponding 3D trajectories, $\{\xi_g\}$, serving as the basis for guiding the robot’s actions. We can also generate more than one trajectory from the same pair of sketches by adding controlled noise to the latent representation.
% However, to address potential issues with trajectory scale or transformation, we implement a calibration step. This step involves using known reference points in the workspace to compute a transformation matrix, which we then apply to the generated trajectories. This ensures that the trajectories are properly scaled and aligned with the actual robot workspace.
% \PRT{This needs to be explained in more details. Either here or in the appendix.}
% With the calibrated 3D trajectories, $\xi_c$, in hand, 
Then we proceed to collect demonstrations for manipulation policy learning. We execute these trajectories on the robotic arm using open-loop servoing, which enables precise trajectory following based on pre-computed motor commands. During execution, we record a demonstration dataset $\{\xi_D = \{(p_t, o_t, a_t)\}_{t=1}^T\}$ at a fixed frequency, where $p_t=(x, y, z)_t$ denotes the robot's end-effector 3D position, $o_t$ represents the robot's observation, $a_t$ is the corresponding action, and $T$ is the total number of timesteps per demonstration. The collected demonstrations, which do not need to be optimal, follow the intended path while capturing the robot's actual behavior in the target environment. They serve as an effective starting point for bootstrapping the policy learning process, offering initial guidance grounded in the robot's real-world performance.

\begin{figure*}
    \centering
    \includegraphics[width=0.85\linewidth]{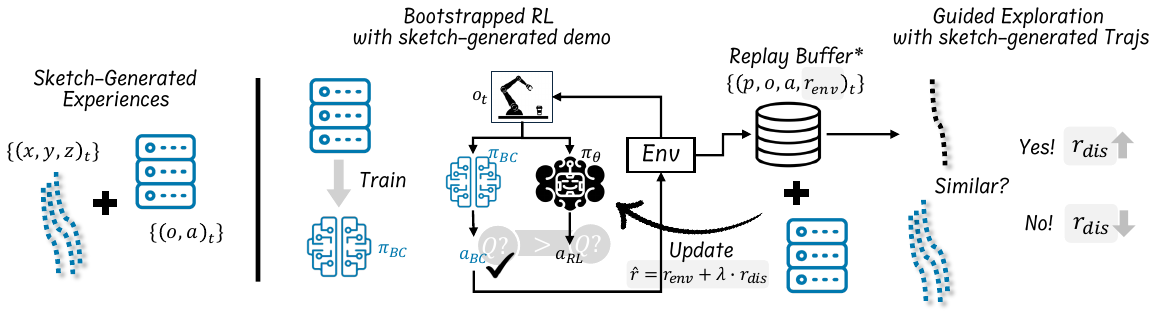}
    \caption{Overview of \OUR{} integrating sketch-generated demonstrations with reinforcement learning. Sketch-generated experiences train an IL policy, which bootstraps the RL process. A discriminator guides exploration by rewarding similarity to sketch-generated trajectories. The final action, combining IL and RL policy outputs, further enhances the exploration guidance. {The asterisk after "Replay Buffer" indicates that the buffer is initialized with the open-loop servoing demonstrations.} }
    \label{fig:policy}
\end{figure*}

\subsection{Policy Learning}
% We now describe our Sketch-To-Skill algorithm (given in Algorithm \ref{alg:sketch_to_skill}).
% The algorithm takes as input the demonstration data $\{\xi_D\}$ collected from our Sketch-to-3D Trajectory Generator $\mT$ and through open-loop servoing (Line 2).
We now describe the policy learning of the \OUR{} algorithm (given in Algorithm \ref{alg:sketch_to_skill}). Taking as input the demonstration data $\{\xi_D\}$ collected from our Sketch-to-3D Trajectory Generator $\mT$ and through open-loop serving (lines 4--5), our approach combines IL and RL to effectively bootstrap and refine the policy. Specifically, we build upon the Imitation Bootstrapped Reinforcement Learning (IBRL) framework \citep{hu2023imitation}, integrating our sketch-based trajectories to guide and constraint policy search space. 
% We do so in two ways. First, we treat the sketch generated trajectories as (potentially sub-optimal) demonstrations. We use an IL policy $\pi_{IL}$ trained on these demonstrations to bootstrap learning of the RL policy $\pi_\theta$. Second, in the actual learning of the RL policy, we use the generated trajectories as additional regularization to help narrow the exploration. 

\begin{algorithm}
\renewcommand{\algorithmiccomment}[1]{#1}
\caption{\OUR{}. Major modifications of IBRL highlighted in \textcolor{RoyalBlue}{blue}.}
\label{alg:sketch_to_skill}
\begin{algorithmic}[1]
\STATE \textbf{Hyperparameters:} Number of critics $E$, number of critic updates $G$, update frequency $U$, exploration std $\sigma$, noise clip $c$, \color{RoyalBlue} number of generated trajectories $m$ per input sketch pair,  reward weighting term $\lambda$
\color{RoyalBlue}\STATE \textbf{Inputs: } Pre-trained Sketch-to-3D Trajectory Generator $\mT$, sketch dataset $\gS = \{(I_i^1, I_i^2)\}_{i=1}^n$,  
\STATE \textbf{Outputs: }\textcolor{RoyalBlue}  Policy $\pi_\theta$, discriminator $D_\psi$\\
% \hfill \COMMENT{\textbf{Stage 1: Demonstration Generation}}
\COMMENT{\makebox[0pt][l]{\rule[0.5ex]{1.2cm}{0.4pt}}\hfill \textit{\textbf{Stage 1}: Demonstration Generation} \hfill \makebox[0pt][r]{\rule[0.5ex]{1.3cm}{0.4pt}}}
% \PRT{can you use better formatting? This can be a comment instead of =====}
% \STATE Generate 3D trajectories $\{\xi_g\}$ from 2D sketches using the pre-trained sketch-to-3D trajectory generator. \PRT{add trained Sketch-to-3D model (use some symbol $T$) in the inputs list. So you can say $\{xi_g\} \leftarrow $ get trajectories from $T$}
\STATE $\{\xi_g\}_{1:mn} \leftarrow $ generate $m$ trajectories per sketch from $\gS$ using $\mT$
% \STATE calibrated trajectories $\xi_c$ \PRT{I would remove this and add this explanation in the appendix. calibration can be thought as just processing of the teacher model or preprocessing of the next statement}
\STATE $\{\xi_D\}_{1:mn} \leftarrow$ generated demonstrations through open-loop servoing\\ 
% \PRT{rollout? this is also where you mention gen3 etc. }

% \hfill\COMMENT{\textbf{Stage 2: Policy Learning}}
\COMMENT{\makebox[0pt][l]{\rule[0.5ex]{2.0cm}{0.4pt}}\hfill \textit{\textbf{Stage 2}: Policy Learning} \hfill \makebox[0pt][r]{\rule[0.5ex]{2.0cm}{0.4pt}}}
\color{black}\STATE Train imitation policy $\pi_{IL}$ on demonstrations \color{RoyalBlue}{$\{\xi_D\}_{1:mn}$} \color{black} using the selected IL algorithm.
\color{black}\STATE Initialize policy $\pi_\theta$, target policy $\pi_{\theta'}$, and critics $Q_\phi$, target critics $Q_{\phi'}$, \textcolor{RoyalBlue}{discriminator $D_\psi$} for $i = 1, \dots, E$
\STATE Initialize replay buffer $B$ with demonstrations \color{RoyalBlue}{$\{\xi_D\}_{1:mn}$} \color{black}
\FOR{$t = 1$ \TO $N$}  % Changed num_rl_steps to N
    \STATE Observe current observation $o_t$ from the environment
    \color{black}\STATE Compute IL action $a^{\text{IL}}_t \sim \pi_{IL}(o_t)$ and RL action $a^{\text{RL}}_t = \pi_\theta(o_t) + \epsilon$, where $\epsilon \sim N(0, \sigma^2)$
    \STATE Sample a set $K$ of 2 indices from $\{1, 2, \dots, E\}$
    \STATE Select action $a_t$ with higher Q-value from $\{a^{\text{IL}}, a^{\text{RL}}\}$
    \STATE Execute action $a_t$
    \STATE Store transition $(\textcolor{RoyalBlue}{p_t}, o_t, a_t, r_t, \textcolor{RoyalBlue}{p_{t+1}}, o_{t+1})$ in replay buffer $B$
    \IF{$t \% U = 0$}  % Changed the condition check
        \color{RoyalBlue}\STATE Perform discriminator $D_\psi$ update by optimizing Equation \ref{eq:discrim} \color{black}
        \STATE Perform TD3 update using minibatches from replay buffer $B$ \color{RoyalBlue} with augmented reward by Equation \ref{eq:reward} \color{black} \citep{fujimoto2018addressing}
    \ENDIF
\ENDFOR
\end{algorithmic}
\end{algorithm}
% \FloatBarrier

% \subsection{Using Sketches during RL}

% \yuph{Amisha: How to present the BC + IBRL algorithm}

% \anu{please check ibrl sketch part below}
% \PRT{Can you refer to the specific lines in the algorithm as part of the text itself? Like.. "Initially, these sketch-based emonstrations are used to train an IL policy (Line 6). }
In IBRL we replace traditional real-world demonstrations with sketch-generated demonstrations. Initially, these sketch-based demonstrations are used to train an IL policy (line 6), which serves as a coarse approximation of the task. Although these sketches do not capture every fine detail of manipulation (e.g., gripper closing/opening actions or exact force control), our hypothesis posits that they still carry significant, actionable information that can effectively guide the learning process in reinforcement learning (RL). We leverage this information
in RL in two ways (as shown in Fig. \ref{fig:policy}):
\begin{enumerate}[label=(\arabic*), leftmargin=1cm, labelsep=0.2cm, itemsep=0pt] 
    \item Bootstrap RL with Sketch-Generated Demos: Even though sketch-generated trajectories are not as detailed as teleoperated demonstrations, they provide a foundational blueprint of the task. We leverage these initial trajectories to bootstrap our RL algorithm, giving it a preliminary direction and reducing the cold start problem common in RL scenarios. This use of imperfect demonstrations is intended to establish an initial policy that avoids random exploration at the outset, making subsequent training more focused and efficient.
    \item Guide Exploration During RL: As the agent progresses in its learning, the sketch-generated trajectories continue to serve as a guide, shaping the exploration strategy. Instead of relying on these trajectories as definitive guides, we treat them as rough outlines that suggest areas of the task space worth exploring. This guided exploration helps concentrate the agent’s learning efforts on potentially fruitful regions of the action space, thus optimizing the learning speed and improving the relevance of the experiences gathered.
\end{enumerate}

In both steps, the use of sketch-generated trajectories acknowledges their limitations—they are not treated as ground truth but as valuable signals to help bootstrap RL and guide exploration throughout the learning process. 
For the RL algorithm, we employ TD3 \citep{fujimoto2018addressing}, an off-policy algorithm known for its sample efficiency. In our approach, the replay buffer is initialized with the sketch-generated demonstration trajectories (line 8), which provide an initial foundation for learning and is later updated with online experiences as the agent interacts with the environment. This combination allows the agent to refine its policy through both sketch-generated demonstration data $\xi_D$ and real-world interaction (line 13).

% \PRT{We need to give some intuition to the algo. Sketch generated demos are not as high quality as actual teleop demos. Plus, they have partial observability as they don't capture gripper motions or fine-grained manipulation strategies. So we can't just use them directly with IL. But our hypothesis is that they still contain useful signal that can guide RL and reduce learning time. We do it in two ways...
% 1. use sketch generated demos (even though they are imperfect) to bootstrap RL
% 2. use sketch generated trajs to guide exploration during RL
% In both ideas, we don't rely on sketch generated trajs are being ground truth.. just as a signal to help start off exploration better in RL (step 1) and guide exploration during RL (step 2). 
% You can have paragraph then explaining each step}

\begin{figure*}[htbp]
    \centering
    \includegraphics[width=0.85\linewidth]{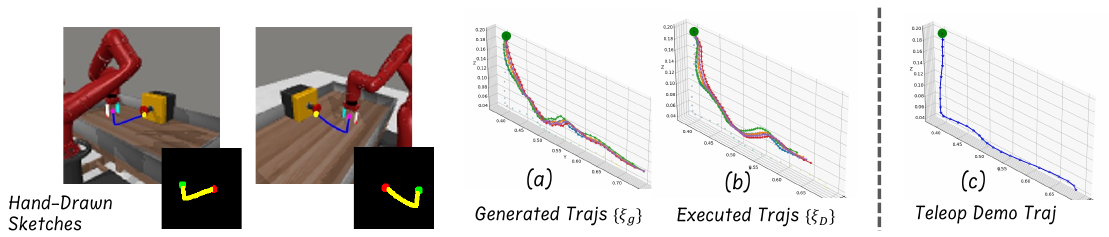}
    \caption{Multi-stage trajectory generation and execution. On the left, we show hand-drawn sketches on scenario RGB images and the extracted sketches on a blank background,  (a) generated trajectory from the Sketch-to-3D Trajectory Generator, and (b) executed trajectory via open-loop serving. In (c), we visualize a teleoperated demo for the same task for reference.}
    \label{fig:hand-drawn}
\end{figure*}

To further enhance the learning process and maintain consistency with the sketch-generated trajectories, we introduce a discriminator-based guided exploration mechanism \citep{kang2018policy}. 
This discriminator, $D_\psi$, is trained to distinguish between trajectories produced by our Sketch-to-3D Trajectory Generator and those generated by the current policy:
\begin{equation}
\begin{aligned}
    \mathcal{L}_{D(\psi)} = \mathbb{E}_{p, g \sim \{\xi_D\}}& [\log D_\psi(p, \Delta p, g)] + \\ &\mathbb{E}_{p, g \sim \pi_\theta} [\log(1 - D_\psi(p, \Delta p, g))],
\label{eq:discrim}
\end{aligned}
\end{equation}

where $p$ represents the end-effector location, $\Delta p$ is the normalized difference between the current and next end-effector positions, capturing local trajectory characteristics, and $g$ is the task-specific information (e.g., target location). This formulation allows the discriminator to assess trajectory similarity while accounting for task variability.
We then augment the TD3 reward function with an additional term based on the discriminator's output (line 18):
\begin{equation}
\hat{r}(o_t, a_t) = r(o_t, a_t) + \lambda \log D_\psi(p_t, \Delta p_t, g),
\label{eq:reward}
\end{equation}
where $\lambda$ is a hyperparameter controlling the influence of the discriminator.
This augmented reward encourages the policy to explore state-action spaces more likely to produce trajectories similar to those generated from human sketches, potentially leading to faster learning and better performance.
% \PRT{We need a pseudocode either here or in the appendix}

Our overall learning process iterates between TD3 optimization and discriminator training. In each iteration: (1) We update the discriminator using the latest policy-generated trajectories and the original sketch-generated trajectories (line 17). (2) We then update the policy and Q-functions using TD3, with the augmented reward (line 18) and guidance from the frozen IL policy (line 13). 
This iterative process allows the policy to refine its behavior while maintaining similarity to the initial demonstrations derived from human sketches. By combining IL, TD3, and discriminator-based guided exploration, we create a cohesive learning framework that effectively leverages sketch-based demonstrations to accelerate and improve the learning of complex manipulation tasks.
Please see the Appendix for more implementation details and a
complete list of hyper-parameters.
%%%%%%%%%%%%%%%%%%%%%%%%%%%%%%%%%%%%%%%%%%%%%%%%%%%%%%%%%%%%%%%%%%%%%%%%%%%%%%%%%%%%%%%%%%%%%%%%
%%%%%%%%%%%%%%%%%%%%%%%%%%%%%%%%%%%%%%%Peseudocode%%%%%%%%%%%%%%%%%%%%%%%%%%%%%%%%%%%%%%%%%%%%%%
%%%%%%%%%%%%%%%%%%%%%%%%%%%%%%%%%%%%%%%%%%%%%%%%%%%%%%%%%%%%%%%%%%%%%%%%%%%%%%%%%%%%%%%%%%%%%%%%%

\section{Experiments}
We report our evaluation of \OUR{}, focusing on its main components: the Sketch-to-3D Trajectory Generator, the Imitation-Bootstrapped RL Policy learning, and the use of the discriminator. Our experiments address the following key questions:
\begin{enumerate}[label=(\arabic*), leftmargin=1cm, labelsep=0.2cm, itemsep=0pt]
    \item[Q1] How effectively does the Sketch-to-3D Trajectory Generator convert 2D sketches into usable 3D robot trajectories?
    \item[Q2] Can \OUR{} utilize sketch-generated demonstrations to achieve comparable performance to traditional methods using high-quality demonstration data?
    \item[Q3] How do various design choices in \OUR{}, such as the number of generated demonstrations per sketch and the discriminator reward weighting, affect the learning and refinement of robotic policy?
    \item[Q4] How well does our method translate to the real world?
\end{enumerate}

\subsection{Evaluation of the Sketch-to-3D Trajectory Generator}
The Sketch-to-3D Trajectory Generator is a key component of \OUR{}, translating 2D sketch inputs into 3D robot trajectories. To train this generator, we collect data of the robot arm executing \textit{play} trajectories. We record the 3D trajectories as well as their 2D projections from two viewpoints. We create such a dataset in the Metaworld \citep{yu2019meta} simulation environment as well as a separate one using actual hardware (Figure~\ref{fig:exp}). Once the Sketch-to-3D Trajectory Generator is trained, we can use hand-drawn sketches as input to predict 3D trajectories.

\textbf{Performance on Hand-drawn Sketches.} We provide an example using the \ttt{ButtonPress} task to qualitatively assess the generator's effectiveness with hand-drawn inputs  (Figure \ref{fig:hand-drawn}). We asked users to provide sketches for the task and also separately collected actual demonstrations as a reference. We see that the Sketch-to-3D Trajectory Generator was able to predict trajectories (Figure \ref{fig:hand-drawn}a) similar to the actual demonstrations (Figure \ref{fig:hand-drawn}c).
We also generate more than one trajectory from the same pair of sketches by adding controlled noise to the latent representation. 
This approach allows us to produce a range of plausible trajectories for a given sketch input, enriching the demonstration set and potentially leading to more robust and adaptable robot policies. We then execute the generated trajectories to produce demonstrations for training the policy (Figure \ref{fig:hand-drawn}b). Despite the inherent variability in sketch inputs, the executed trajectory further validates the practical applicability of our approach. This demonstrates our model's robustness to sketch imperfections and its ability to reliably interpret user intent, bridging the gap between simple 2D sketches and actionable 3D robot trajectories.

\textbf{Latent Space Representation and Interpolation.} To further understand the generator's latent space, we performed linear interpolation in the latent space between different input samples. Specifically, we selected two distinct sketch pairs with different trajectories, extracted their feature vectors, linearly interpolated between them, reconstructed the sketches, and generated new trajectories.
Figure \ref{fig:interpolated} shows smooth transitions in both 2D sketches and 3D trajectories across the interpolated latent space. This smoothness demonstrates that our model has learned a continuous and semantically meaningful representation, suggesting good generalization capability to unseen inputs that lie between known examples \citep{kingma2013auto}. The coherence between interpolated sketches and their corresponding 3D trajectories further validates the model's robust sketch-to-trajectory mapping.
% \PRT{is there a reference to another work that uses a similar visualization to demonstrate the goodness of a model? if so, would be nice to cite here saying we are following the same idea.}

\textbf{Effect of VAE in Sketch-to-3D Trajectory Generator.}
We also conducted an ablation study of the Sketch-to-3D Trajectory Generator to evaluate the effect of the VAE on the architecture. Using a dataset of 1000 trajectories split 80:20 for training and validation, we report the training and validation losses in Table~\ref{tab:vae}. 
We compare the performance of the model with the VAE (shown in Figure \ref{fig:teacher}) to a variant without the VAE, where only a CNN is used. This CNN has the same architecture as the VAE's encoder, but without the decoder and loss components. Our results show that incorporating the VAE consistently improves performance across all metrics, including reconstruction loss and trajectory loss.

\begin{table}[ht]
\centering
\caption{Performance Metrics for generator model}
\label{tab:vae}
\begin{tabular}{
  @{}l 
  >{\raggedright\arraybackslash}p{3.0cm} 
  >{\raggedright\arraybackslash}p{3.0cm} 
  >{\raggedright\arraybackslash}p{3.0cm} 
  >{\raggedright\arraybackslash}p{3.0cm}@{}
}
\toprule
Metric  & with VAE & without VAE \\ \midrule
Training Loss          & \textbf{0.6019 }& 0.6286\\
Validation Loss        & \textbf{0.9128} & 1.2804\\
Reconstruction Loss   & \textbf{0.0004} & 0.3258 \\
KLD Loss              & \textbf{69.8592} & 107.167 \\
Parameter MSE Loss    & \textbf{0.0820} & 0.0928 \\
Trajectory Loss    & \textbf{0.770} & 0.1180 \\ \bottomrule
\end{tabular}
\end{table}

\subsection{Comparisons with Baselines} In this section, we conduct extensive experiments in Robomimic \cite{robomimic2021} and MetaWorld \citep{yu2019meta} to answer Q2: \emph{can \OUR{} utilize sketch-generated demonstrations to achieve comparable performance to traditional methods using high-quality demonstration data?}
Specifically, we compare \OUR{} with: (1) IBRL \citep{hu2023imitation}, a strong baseline that utilizes traditional high-quality demonstration data (rather than sketches as what our method uses), and (2) TD3 \citep{fujimoto2018addressing}, a state-of-the-art pure RL approach without using any demonstrations. We hypothesize that although the sketches have only partial information (namely, 2D projections of 3D trajectories and no gripper information), we can still generate good enough demonstration data to perform comparably with the baseline that uses full demonstrations. We show that to be the case in these experiments.

% As baselines we use (1) Imitation-Based RL (IBRL) with teleoperated demonstrations and (2) Pure RL using the environment reward. We also compare \OUR{} with and without the discriminator reward. Figures \ref{fig:rl_train_comparison} and \ref{fig:rl_eval_comparison} show the training and evaluation performance across all the tasks.
% \PRT{remember to explain what the scores mean (success rate? returns?) in the figure captions} 

We perform evaluations on two-stage \ttt{PickPlaceCan} task from Robomimic and six tasks from the MetaWorld benchmark, namely \ttt{Coffeepush}, \ttt{Boxclose}, \ttt{Buttonpress}, \ttt{Reach}, \ttt{Reachwall}, and \ttt{ButtonpressTopdownwall}, each using sparse 0/1 task completion rewards at the end of each episode. 
% These tasks cover easy and hard difficulty levels as categorized in \cite{seo2023masked}.
For each task, we collected 10 high-quality demonstrations using an expert policy in Robomimic and 3 high-quality demonstrations in MetaWorld. These demonstrations served as our baseline for traditional demonstration-based methods. {For our approach, we collected a total of three hand-drawn sketches, one on each demonstration's initial frames (Figure \ref{fig:hand-drawn})}. These sketches were used to generate and execute trajectories, creating a parallel set of sketch-based demonstrations for comparison.

\begin{figure}[h!]
    \centering
    \begin{subfigure}[c]{\columnwidth}
        \includegraphics[width=\linewidth]{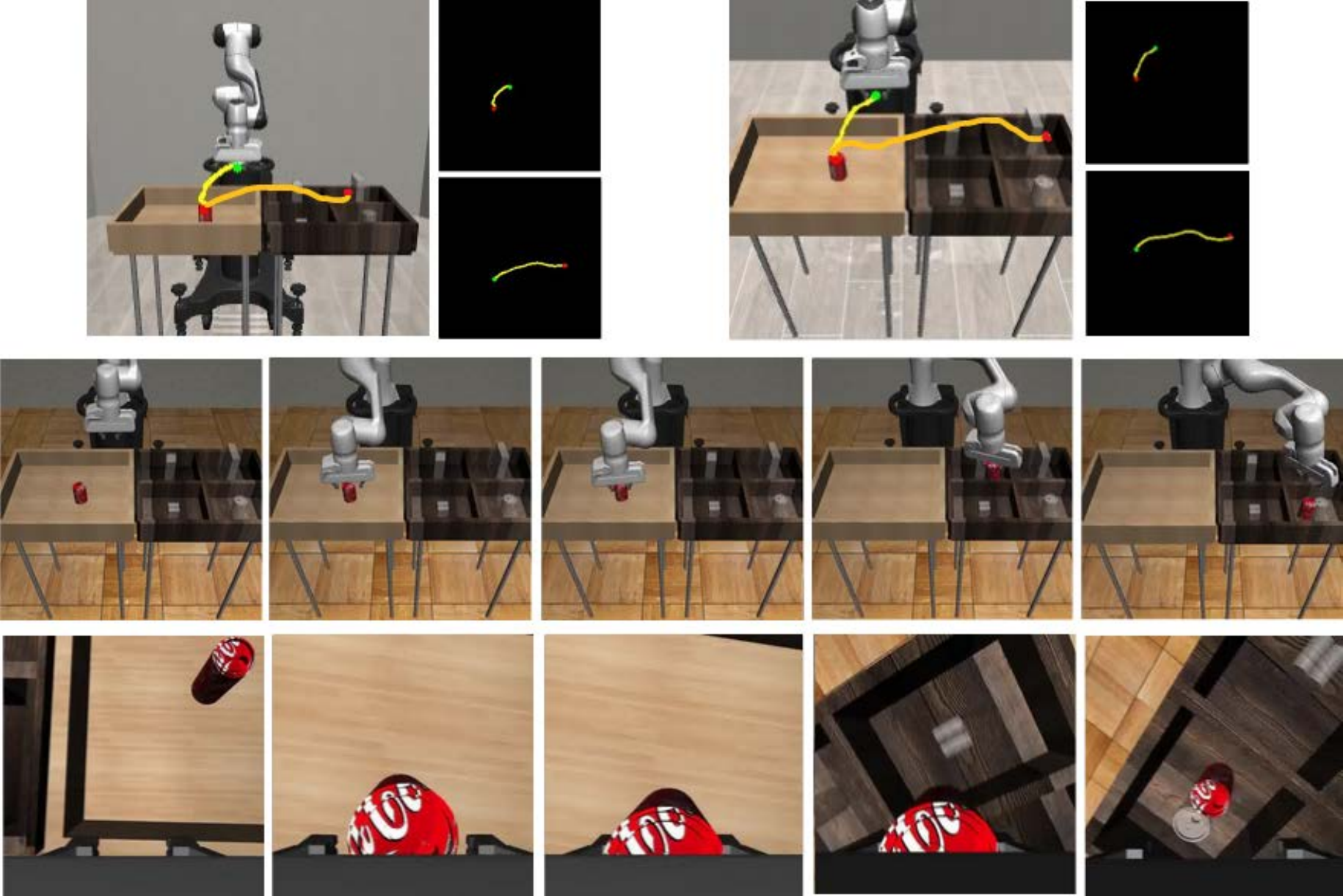}
    \end{subfigure}
    \vspace{1em} % Optional spacing between the two subfigures
    \vspace{0.1em}
    \begin{subfigure}[c]{\columnwidth}
    \includegraphics[width=\linewidth]{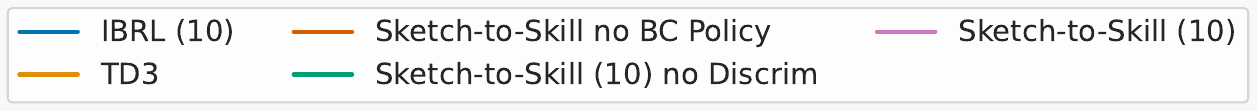}
    \centering
    \includegraphics[width=0.8\linewidth]{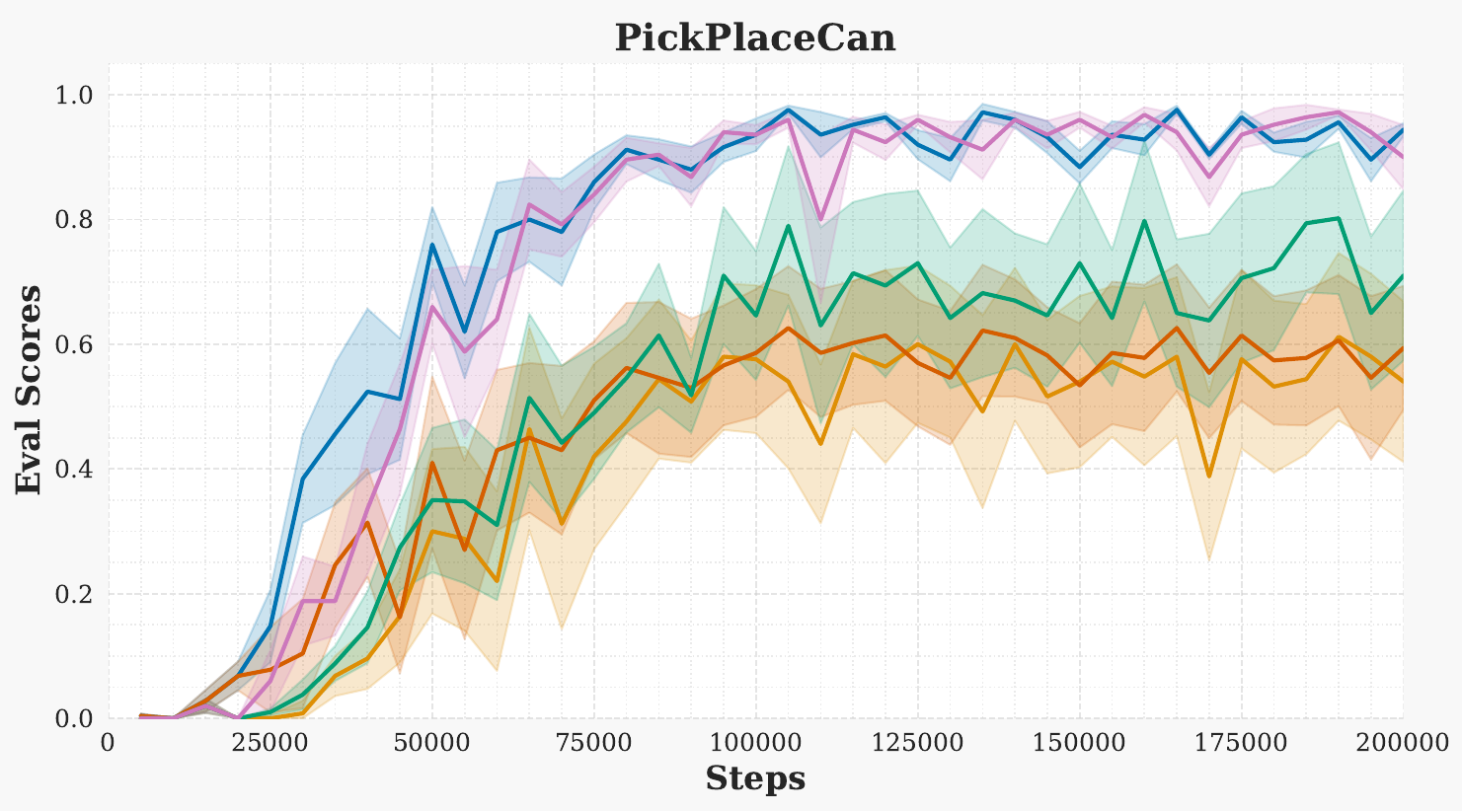}
    \end{subfigure}
    \caption{Evaluation Scores (success rate) for the robomimic PickPlaceCan environment during evaluation.}
    \label{fig:pick-place-can}
\end{figure}

\begin{figure*}[t]
\centering
%%%%%%%%%% original version
% \begin{subfigure}[c]{0.3\columnwidth}
%     \centering
%      \includegraphics[width=0.88\textwidth]{figures/rl_comparison/RL_eval_comparison_ButtonPress.pdf}
%      % \caption{ButtonPress}
%      \label{fig:rl_eval_ButtonPress}
% \end{subfigure}
% \begin{subfigure}[c]{0.3\columnwidth}
% \centering
%      \includegraphics[width=0.88\textwidth]{figures/rl_comparison/RL_eval_comparison_CoffeePush.pdf}
%      % \caption{CoffeePush}
%      \label{fig:rl_eval_CoffeePush}
% \end{subfigure}
% \begin{subfigure}[c]{0.3\columnwidth}
% \centering
%      \includegraphics[width=0.88\textwidth]{figures/rl_comparison/RL_eval_comparison_BoxClose.pdf}
%      % \caption{BoxClose}
%      \label{fig:rl_eval_BoxClose}
% \end{subfigure}
% \begin{subfigure}[c]{0.3\columnwidth}
% \centering
%      \includegraphics[width=0.88\textwidth]{figures/rl_comparison/RL_eval_comparison_Reach.pdf}
%      % \caption{Reach}
%      \label{fig:rl_eval_Reach}
% \end{subfigure}
% \begin{subfigure}[c]{0.3\columnwidth}
% \centering
%      \includegraphics[width=0.88\textwidth]{figures/rl_comparison/RL_eval_comparison_ReachWall.pdf}
%      % \caption{ReachWall}
%      \label{fig:rl_eval_ReachWall}
% \end{subfigure}
% \begin{subfigure}[c]{0.3\columnwidth}
% \centering
%      \includegraphics[width=0.88\textwidth]{figures/rl_comparison/RL_eval_comparison_ButtonPressTopdownWall.pdf}
%      % \caption{ButtonPressTopdownWall}
%      \label{fig:rl_eval_ButtonPressTopdownWall}
% \end{subfigure}
%%%%%%%%%%% color-blind friendly version
\includegraphics[width=0.9\textwidth]{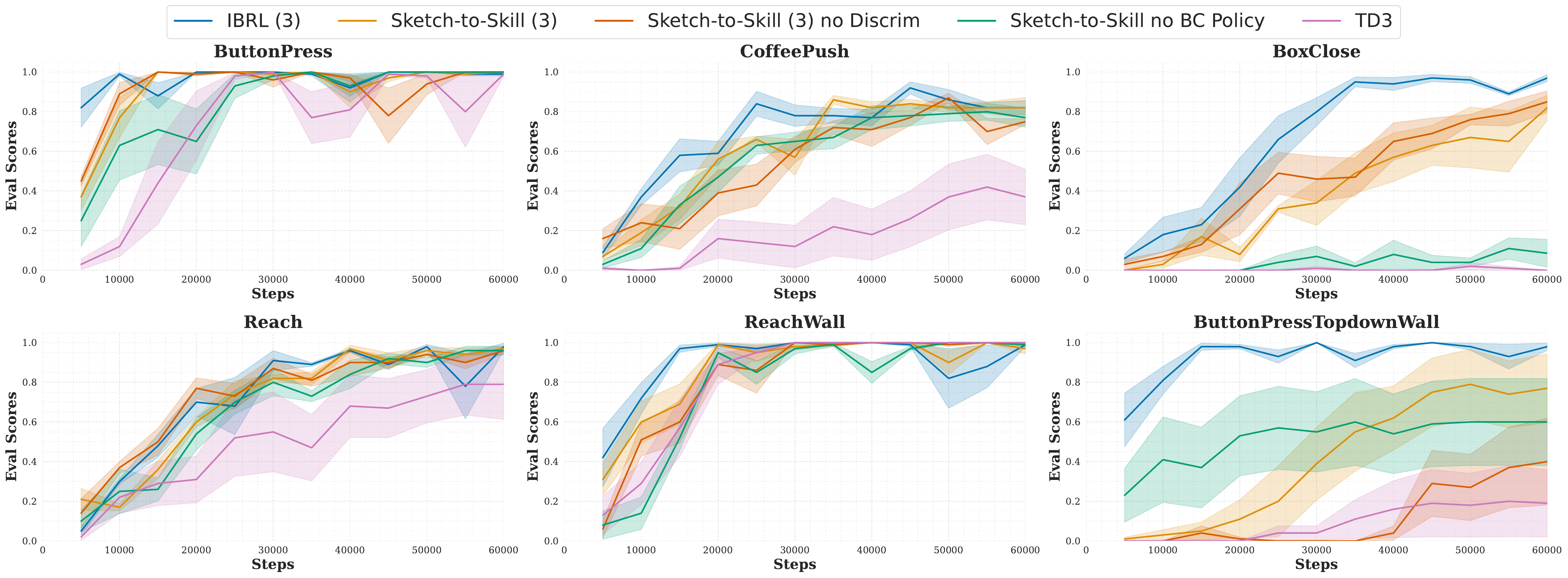}
\caption{ {\textbf{Evaluation Performance of \OUR{} in MetaWorld }This figure shows the success rate across six MetaWorld tasks under randomized initial gripper and object positions. \OUR{} achieves comparable performance to IBRL while surpassing pure RL. }}
\label{fig:rl_eval_comparison}
\end{figure*}

\begin{figure*}
    \centering
    \includegraphics[width=0.85\linewidth]{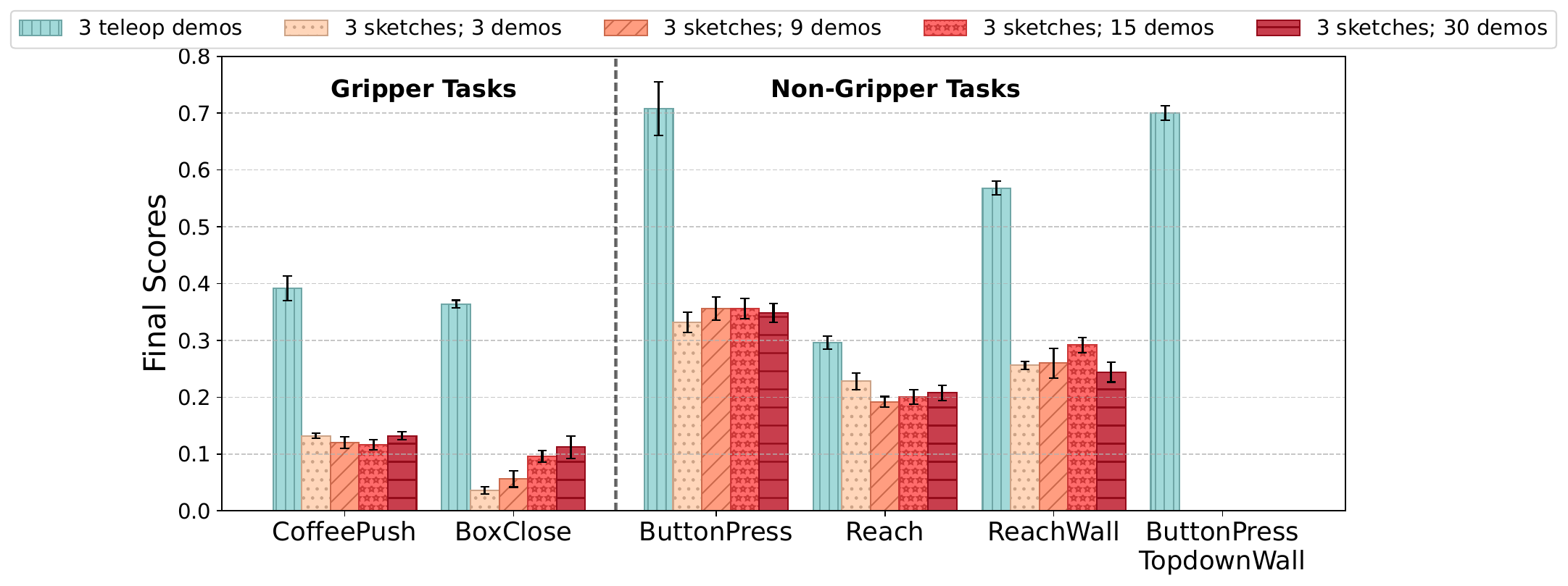}
    \caption{\textbf{Behavioral Cloning (BC) scores using actual teleoperated data and sketch generated demonstrations}. The blue bars represent the baseline BC policy trained with 3 high-quality demonstrations, while the red bars show BC policies trained with sketch-generated demonstrations, varying in the number of demonstrations $m$ per input sketch pair (1, 3, 5, and 10). Darker shades of red indicate an increase in the number of sketch-based demonstrations used for training. Despite poor success rate, the actual trajectories and policy learned with sketches are useful for bootstrapping as evidenced by the training performance (Figures~\ref{fig:rl_eval_comparison}).
    % This visualization highlights the dependency of BC performance on the quality and quantity of the training demonstrations across multiple tasks.
    }
    \label{fig:bc_comparison}
\end{figure*}

Figure \ref{fig:pick-place-can} shows the setup where the sketch-based demonstrations provide initial positional cues for the robot in the Robomimic environment. We ask users to draw separate sketches for each stage, and then combine the trajectories during open-loop servoing.  Our framework utilizes RL to refine these initial cues, dynamically adjusting the robot’s approach to manage both the orientation and timing necessary for successful task execution. This method proves particularly effective as it does not rely on fully detailed trajectory information from the start; instead, it uses sketches to guide the initial exploration phase of RL, simplifying the data collection process. Our sketch-to-skill framework, even with limited initial data, performs admirably in this demanding scenario. The results are on par with those from IBRL methods $\sim$98$\%$, which benefit from complete and detailed human demonstrations, including explicit orientation details (as shown in Figure 5).

% Figures \ref{fig:rl_train_comparison} and \ref{fig:rl_eval_comparison} show the training and evaluation performance across all the tasks in the MetaWorld environment. 
% We present \OUR{}'s results with and without the discriminator reward. 
Figures \ref{fig:rl_eval_comparison} show the evaluation performance across all the tasks in the MetaWorld environment. 
To assess the effectiveness of individual components, we also compare \OUR{}’s results with variants where the BC policy guidance or the discriminator reward  is removed.
% We observe that in all cases \OUR{} performs better, often significantly better than pure RL. In most tasks, \OUR's performance using only sketches as input is comparable to IBRL which uses high-quality demonstrations. This is particularly notable in the \ttt{CoffeePush} and \ttt{Boxclose} tasks.
We observe that \OUR{} consistently outperforms pure RL across all tasks, often by a notable margin. Both the BC policy guidance and the discriminator reward contribute to policy learning, resulting in the most stable performance across tasks. 
Remarkably, \OUR{} achieves performance comparable to IBRL, which relies on high-quality demonstrations, even when using only sketches as input. This is particularly evident in the \ttt{CoffeePush} and \ttt{BoxClose} tasks.
% \PRT{insert names of tasks that uses gripper} 
These tasks require actuating the gripper --- information that is not provided in the sketches. Nevertheless, \OUR{} is able to bootstrap and use guidance from the sketch generated suboptimal demonstrations to learn a policy efficiently. 
This provides evidence to the claim that the sketch-generated demonstrations do not lead to much degradation in performance while being much easier to obtain.

\begin{figure*}[h]
\centering
% \begin{subfigure}[c]{0.7\linewidth}
%     \centering
%      \includegraphics[width=1\textwidth]{figures/legend_no_discrim_red.pdf}
% \end{subfigure}
% % \vspace{-0.2}
% \begin{subfigure}[c]{0.3\linewidth}
%     \centering
%      \includegraphics[width=0.88\textwidth]{figures/rl_comparison/ablation_no_discrim_ButtonPress.pdf}
% \end{subfigure}
% \begin{subfigure}[c]{0.3\linewidth}
% \centering
%      \includegraphics[width=0.88\textwidth]{figures/rl_comparison/ablation_no_discrim_CoffeePush.pdf}
% \end{subfigure}
% \begin{subfigure}[c]{0.3\linewidth}
% \centering
%      \includegraphics[width=0.88\textwidth]{figures/rl_comparison/ablation_no_discrim_BoxClose.pdf}
% \end{subfigure}
% \begin{subfigure}[c]{0.3\columnwidth}
% \centering
%      \includegraphics[width=0.88\textwidth]{figures/rl_comparison/ablation_no_discrim_Reach.pdf}
% \end{subfigure}
% \begin{subfigure}[c]{0.3\columnwidth}
% \centering
%      \includegraphics[width=0.88\textwidth]{figures/rl_comparison/ablation_no_discrim_ReachWall.pdf}
% \end{subfigure}
% \begin{subfigure}[c]{0.3\columnwidth}
% \centering
%      \includegraphics[width=0.88\textwidth]{figures/rl_comparison/ablation_no_discrim_ButtonPressTopdownWall.pdf}
% \end{subfigure}
% \caption{Ablation training scores for \OUR{} without discriminator}
% \label{fig:ablation_no_discrim}
% \end{figure}

% \begin{figure}[h]
\centering
\begin{subfigure}[c]{0.7\linewidth}
    \centering
     \includegraphics[width=1\textwidth]{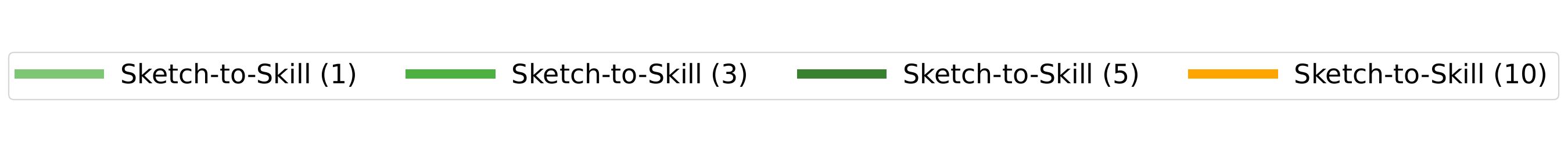}
\end{subfigure}
\begin{subfigure}[c]{0.3\linewidth}
    \centering
     \includegraphics[width=0.88\textwidth]{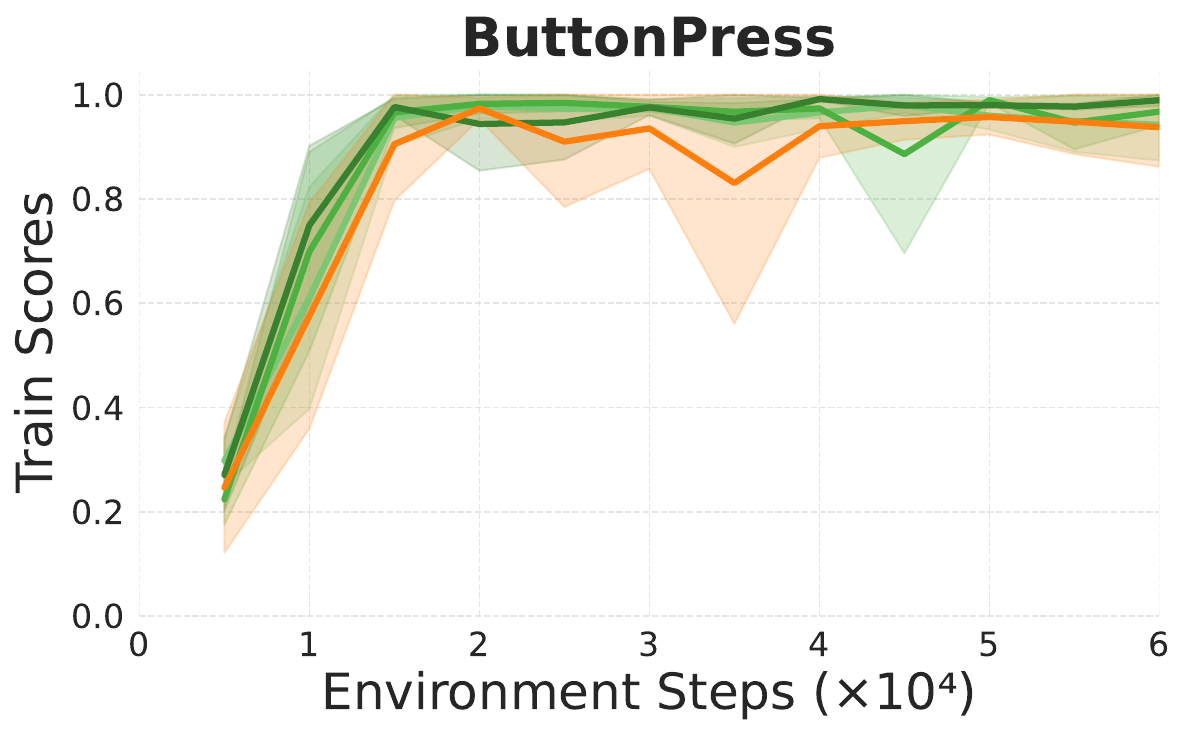}
\end{subfigure}
\begin{subfigure}[c]{0.3\linewidth}
\centering
     \includegraphics[width=0.88\textwidth]{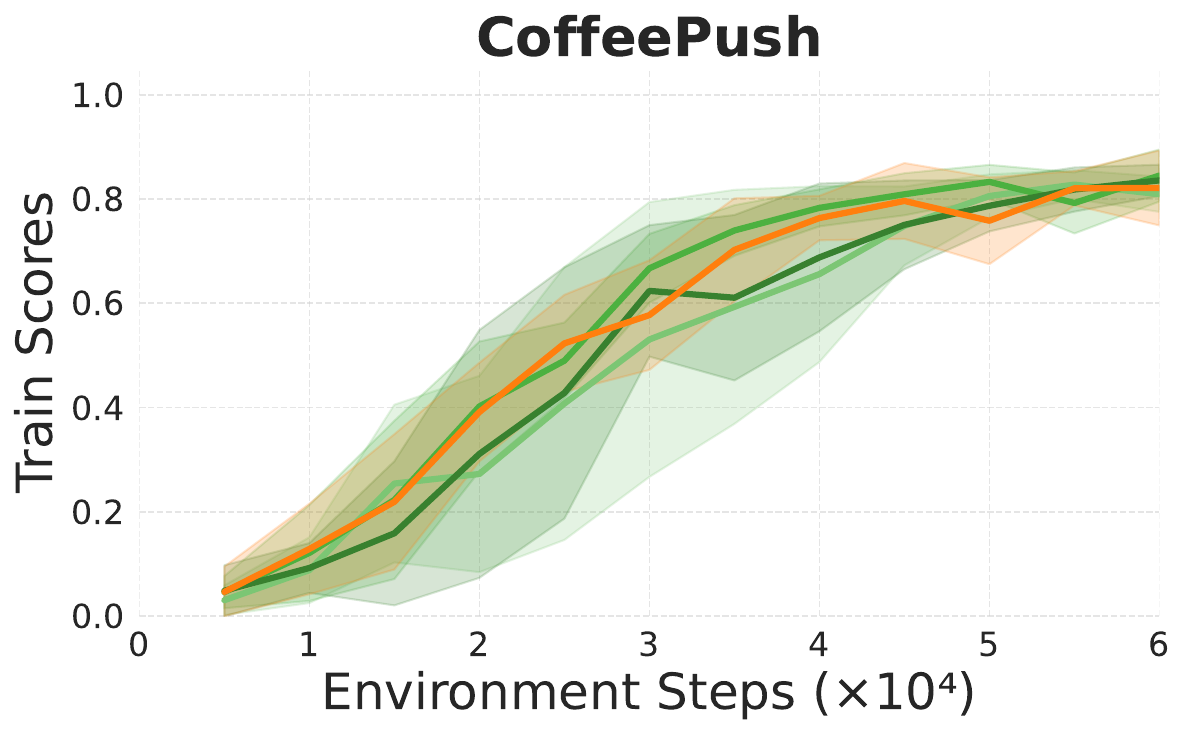}
\end{subfigure}
\begin{subfigure}[c]{0.3\linewidth}
\centering
     \includegraphics[width=0.88\textwidth]{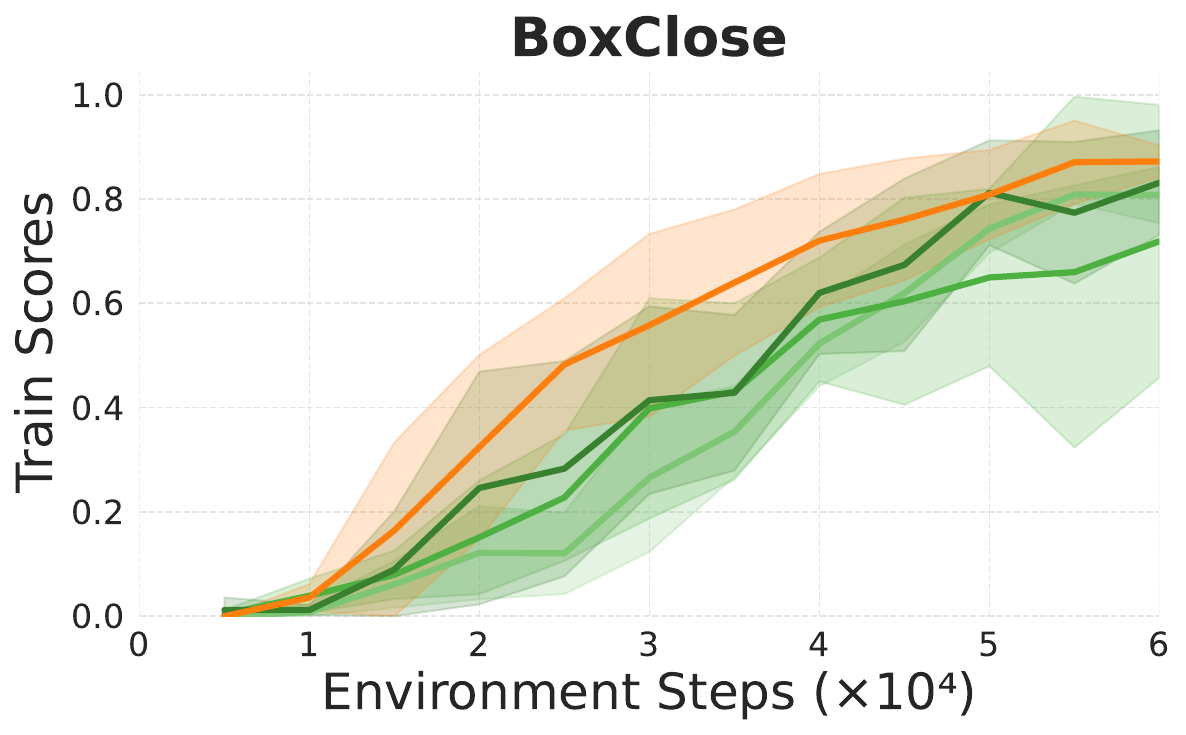}
\end{subfigure}
\begin{subfigure}[c]{0.3\linewidth}
\centering
     \includegraphics[width=0.88\textwidth]{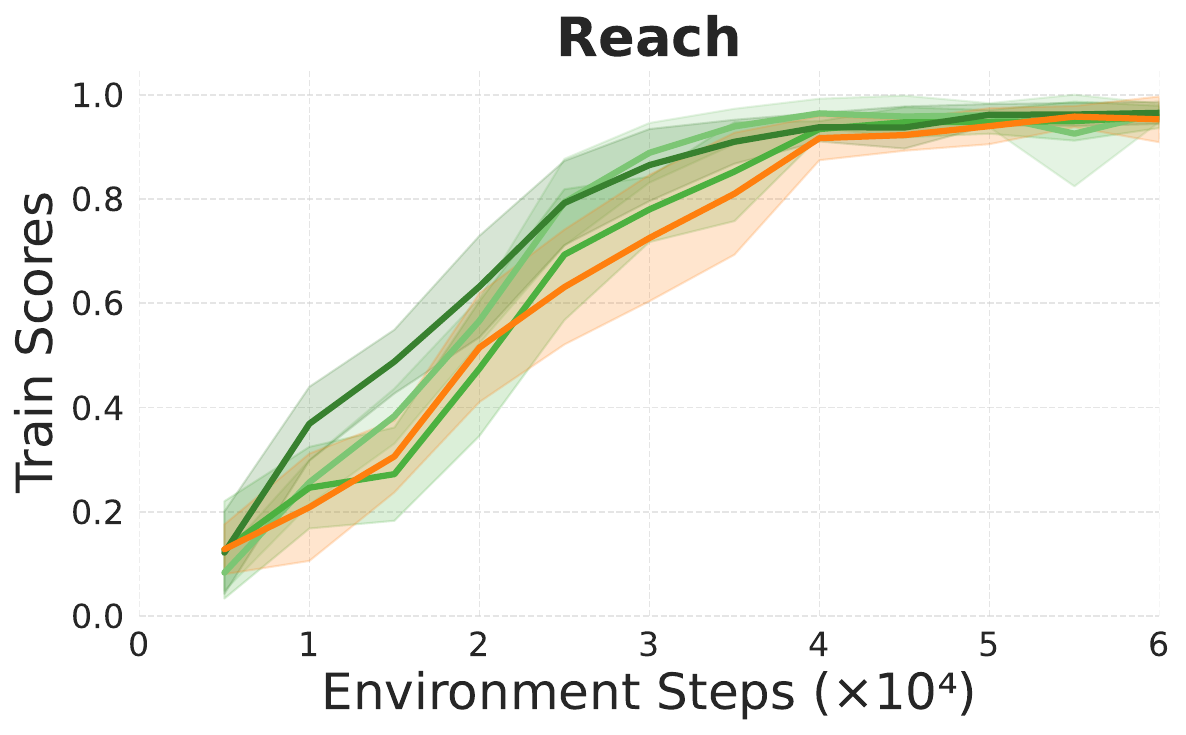}
\end{subfigure}
\begin{subfigure}[c]{0.3\linewidth}
\centering
     \includegraphics[width=0.88\textwidth]{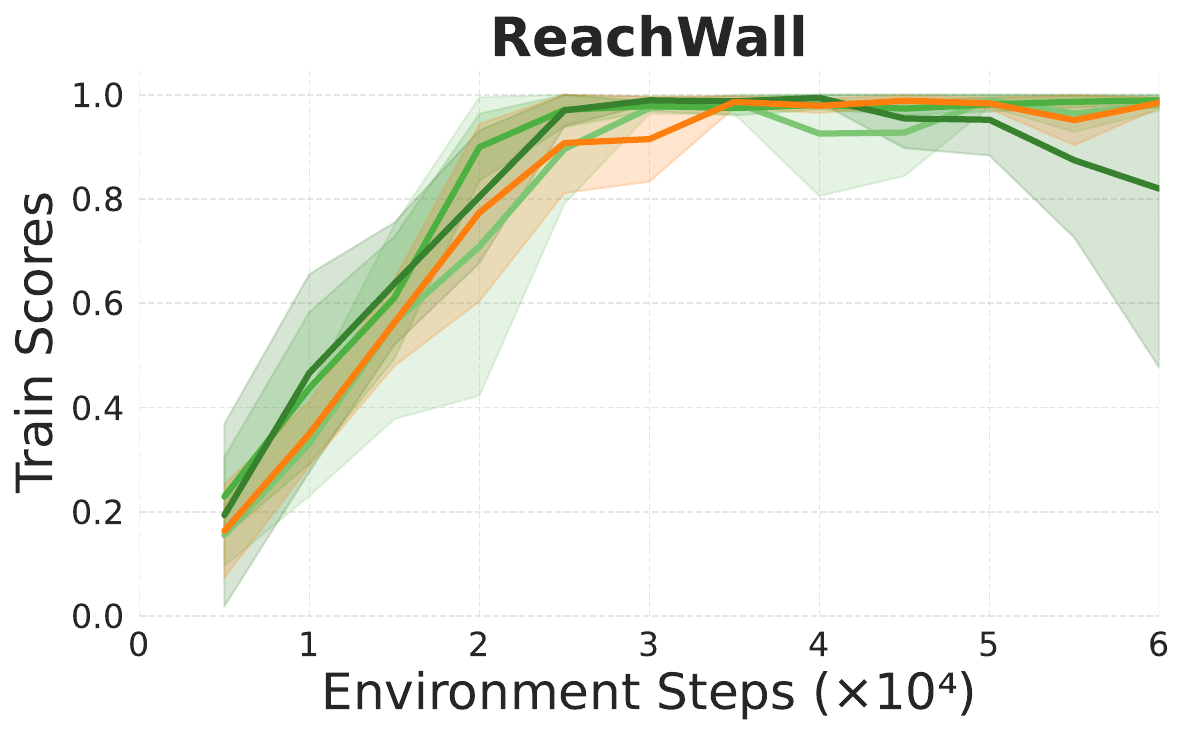}
\end{subfigure}
\begin{subfigure}[c]{0.3\linewidth}
\centering
     \includegraphics[width=0.88\textwidth]{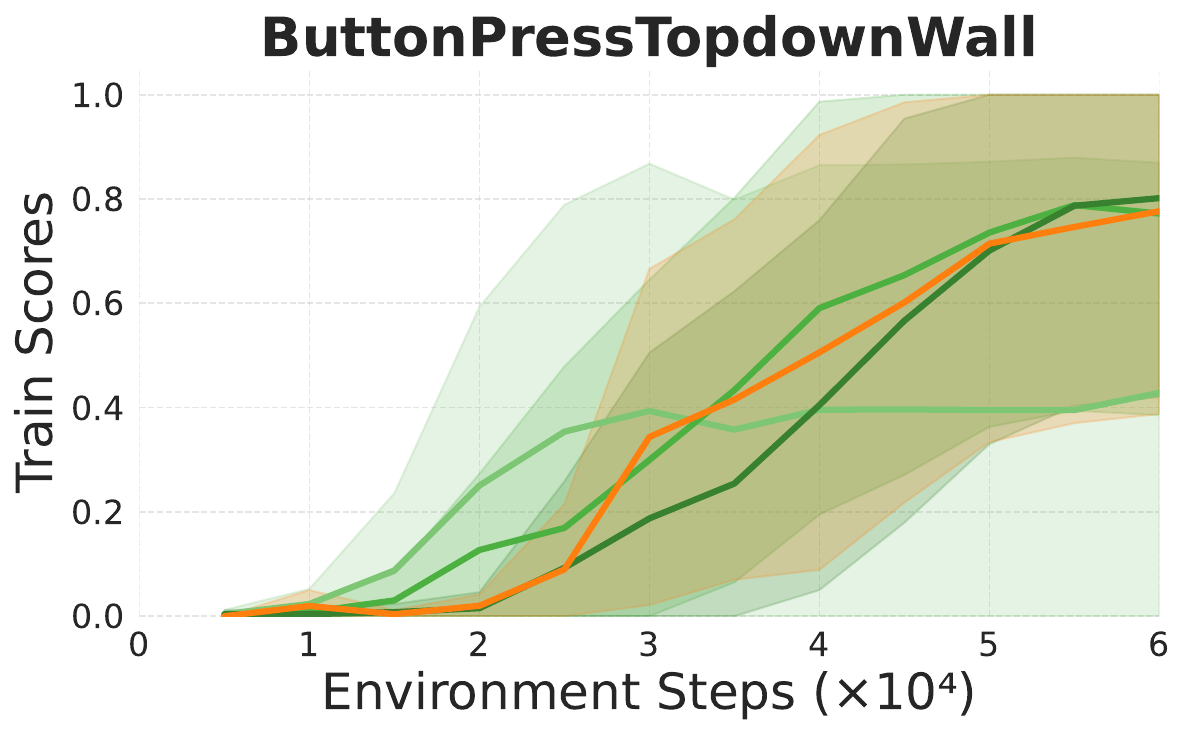}
\end{subfigure}
\caption{
% Top row illustrates ablation training scores for \OUR{} without discriminator and the bottom row shows with discriminator. 
This figure illustrates ablation training scores for \OUR{} with varying $m$, the number of demonstrations generated per sketch pair (1, 3, 5, and 10). }
% \label{fig:ablation_no_discrim }
\label{fig:ablation_sketch_to_skill}
\end{figure*}

\textbf{Behavioral Cloning Performance.} \OUR{} employs behavior cloning (BC) to bootstrap policy learning, similar to IBRL. However, the key difference is that IBRL relies on high-quality teleoperated demonstrations, whereas \OUR{} uses sketch-generated demonstrations. We compare the performance between them (Figure \ref{fig:bc_comparison}) and ablate the number of generated demonstrations $m$ per input sketch pair. 
% \OUR{} uses BC to bootstrap learning of the policy. We also compare the BC performance of policies trained on: (1) teleoperated demonstrations, and (2) sketch-generated demonstrations (Figure \ref{fig:bc_comparison}). The actual BC architecture is the same in both cases, just the input data varies. 
Not surprisingly the BC policy with teleoperated data performs better than the sketch generated ones. However, despite the lower performance of the BC policy, \OUR{} is still able to achieve comparable performance in RL training (as seen in Figures~\ref{fig:pick-place-can} and \ref{fig:rl_eval_comparison}), showing that it is not as sensitive to the quality of the bootstrapping policy. Increasing the number of generated demonstrations $m$ per input sketch pair (from 1 to 10) does not significantly improve the BC performance.

\subsection{Ablation Studies}
To understand the impact of the key components in \OUR{} and answer Q3, we conducted ablation studies focusing on two critical aspects: the number of generated trajectories $m$ per input sketch pair and the reward weighting scheme $\lambda$.

% \PRT{use the symbols here and in the \textbf{bold titles} in the following; I think $m$ and $\lambda$ but verify}

\textbf{Impact of Generated Trajectories per Sketch.} We investigated how the number of trajectories generated from each input sketch pair affects the learning performance. Figure \ref{fig:ablation_sketch_to_skill},  shows the learning curves for policies trained with varying numbers of generated trajectories per sketch. 
% We see that the performance is improved when we generate $m=3$ trajectories per sketch, instead of just one trajectory per sketch. Here, the additional demonstrations can make up for the deficiency of not having actual teleoperated demonstrations. However, increasing the number of trajectories per sketch has diminishing value. It is useful when the tasks are difficult, such as \ttt{BoxClose} and \ttt{CoffeePush} which involve gripper actions, but does not affect much for easier tasks.
We observe improved performance when generating $m=3$ trajectories per sketch instead of just one. The additional demonstrations help compensate for the lack of actual teleoperated demonstrations. However, the benefit of increasing the number of trajectories per sketch becomes less pronounced beyond a certain point. This approach is particularly useful for challenging tasks, such as \ttt{BoxClose} and \ttt{CoffeePush}, which involve precise gripper actions, but has limited impact on simpler tasks.

\begin{figure}[b]
    \centering
    \includegraphics[width=0.75\linewidth]{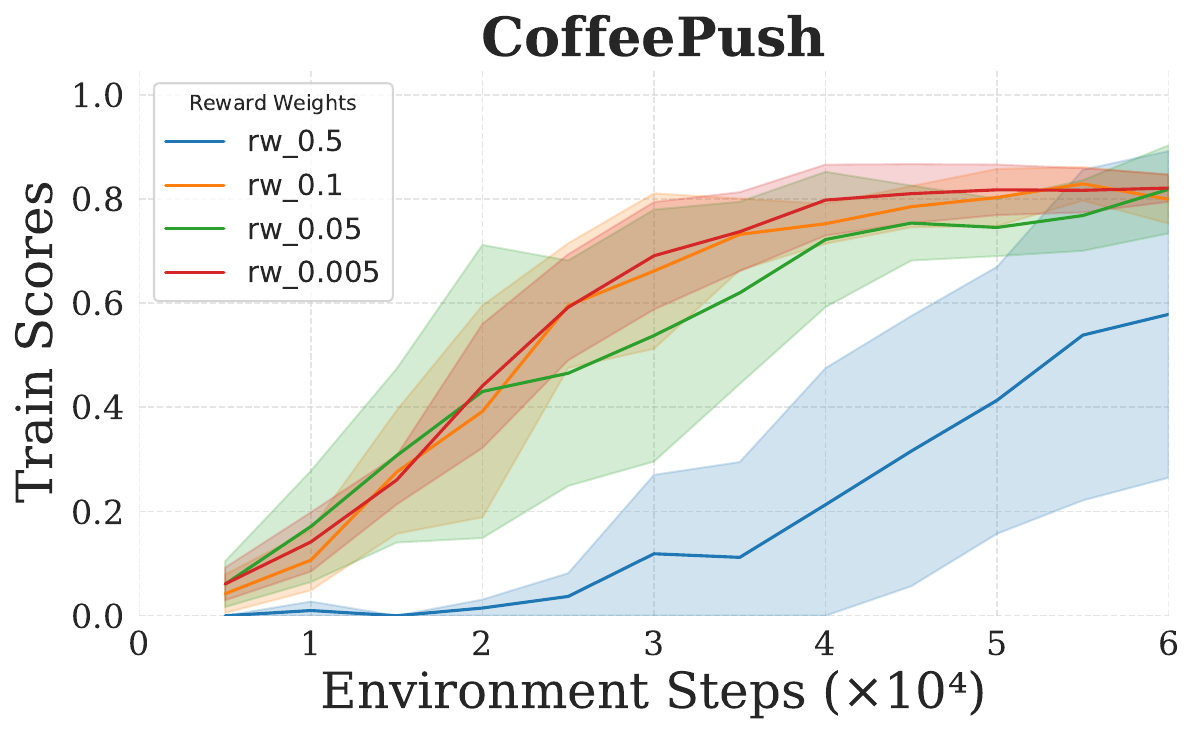}
    \caption{{Reward weighting term ablation}}
    \label{fig:ablation_rw}
\end{figure}

% \PRT{What's the takeaway? We see that the performance is improved when we general $m=3$ trajectories per sketch, instead of just one trajectory per sketch. Here, the additional demonstrations can make up for the deficiency of not having actual teleoperated demonstrations. However, increasing the number of trajectories per sketch has .. I don't know the plot but either say diminishing value if it improves but not by much, or say a detrimental effect if the performance actually goes down. }

\textbf{Effect of Reward Weighting}: We examined the impact of different reward weighting schemes on policy learning. Our reward function combines the environmental reward with a discriminator-based reward by Equation \ref{eq:reward}, where $\lambda$ is the weighting parameter. 
% Figure \ref{fig:ablation_rw} illustrates the learning performance across different values of $\lambda$.The model demonstrates comparable performance with reward weights of 0.1 and 0.005, but significantly underperforms with a reward weight of 0.5. 
Figure \ref{fig:ablation_rw} shows the learning performance across various $\lambda$ values: 0.005, 0.05, 0.1, and 0.5. The model performs comparably with $\lambda$ values of 0.005, 0.05, and 0.1, indicating relative insensitivity to this hyperparameter. However, it significantly underperforms when $\lambda$ is set to 0.5.

% \PRT{need a takeaway as well}
% \PRT{Peihong: organize this around questions like in TMLR submission}

\subsection{Hardware Experiments}
We validate \OUR{} on physical robot hardware to demonstrate its effective transfer from simulation to real-world applications.

\begin{figure}[p]
\centering

\begin{subfigure}[c]{\columnwidth}
    \centering
    \includegraphics[width=0.8\linewidth]{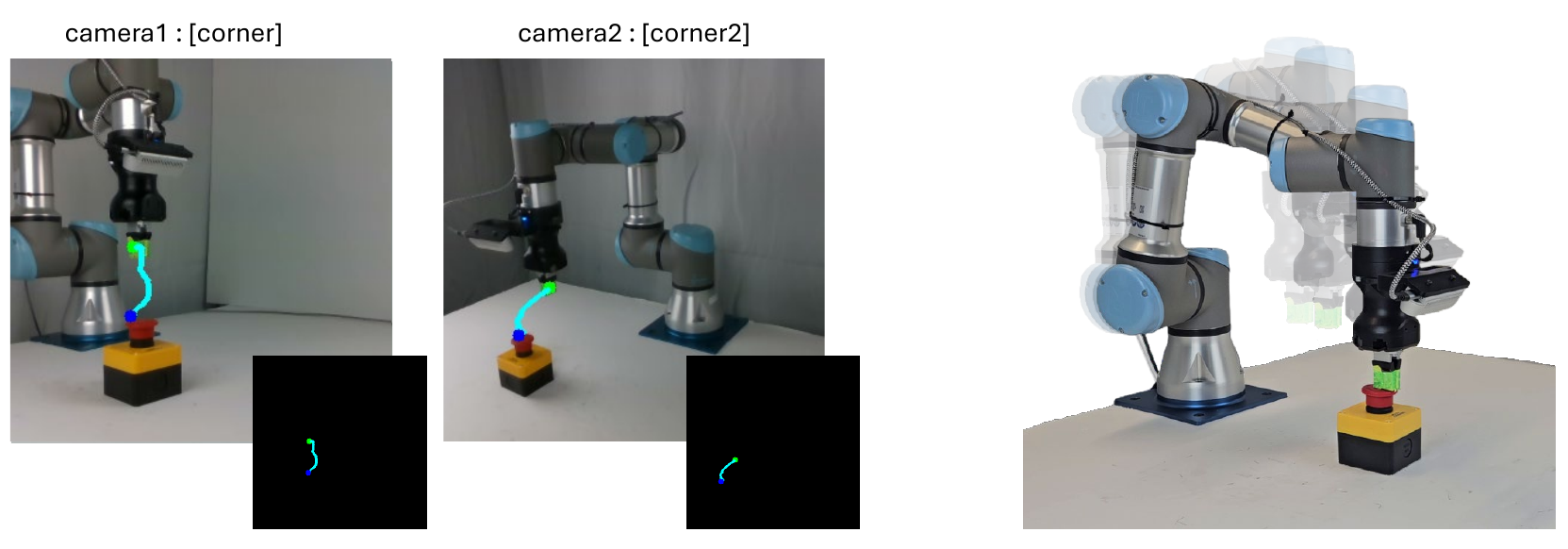}
    % \caption{Hardware setup for the real-world ButtonPress task.}
    \label{fig:sketch_hardware}
\end{subfigure}
\vspace{1em} % Optional spacing
\begin{subfigure}[c]{\columnwidth}
    \centering
    \includegraphics[width=0.6\linewidth]{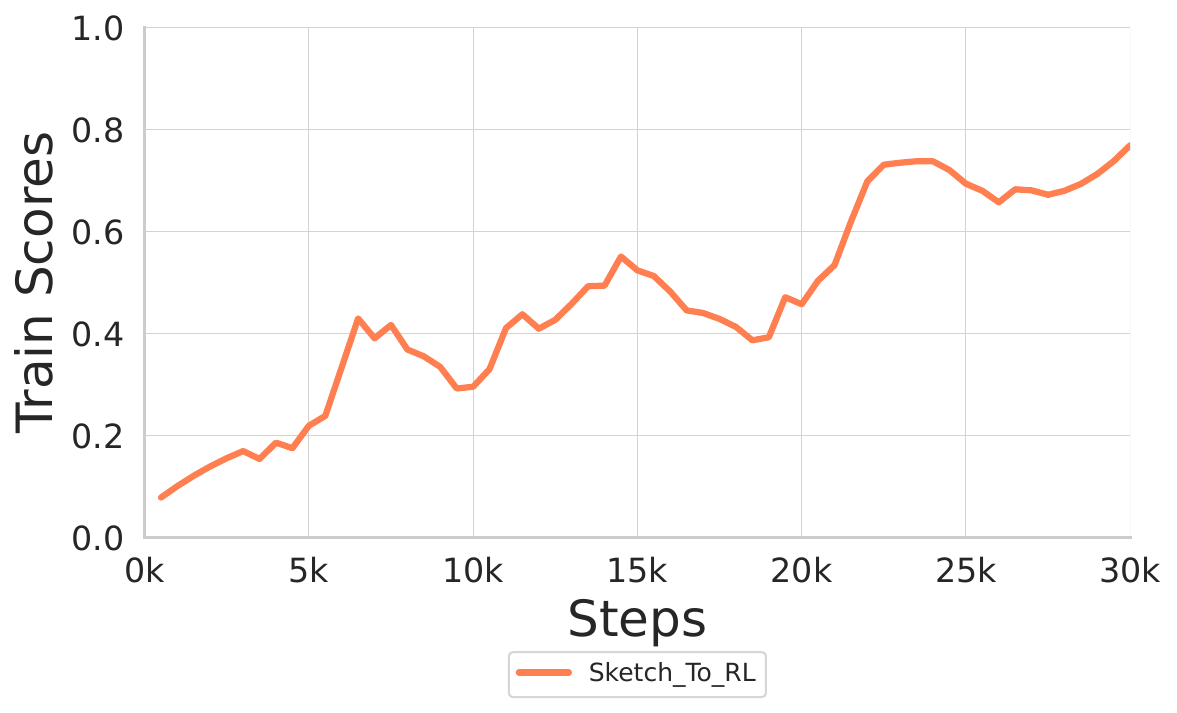}
    \caption{ButtonPress.}
    \label{fig:hardware_scores}
\end{subfigure}

\begin{subfigure}[c]{\columnwidth}
    \centering
    \includegraphics[width=0.8\linewidth]{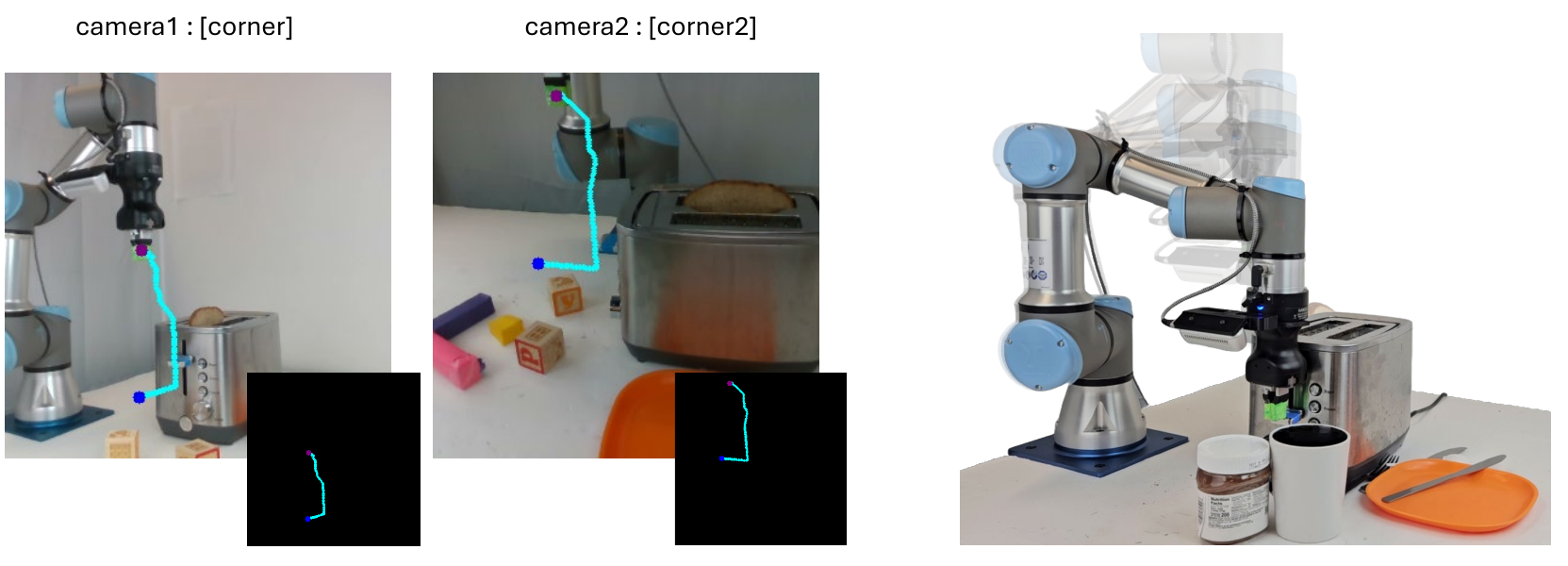}
    % \caption{Training curves of the ToastButtonPress task in a cluttered environment.}
    \label{fig:toastpress}
\end{subfigure}
\vspace{1em} % Optional spacing
\begin{subfigure}[c]{\columnwidth}
    \centering
    \includegraphics[width=0.55\linewidth]{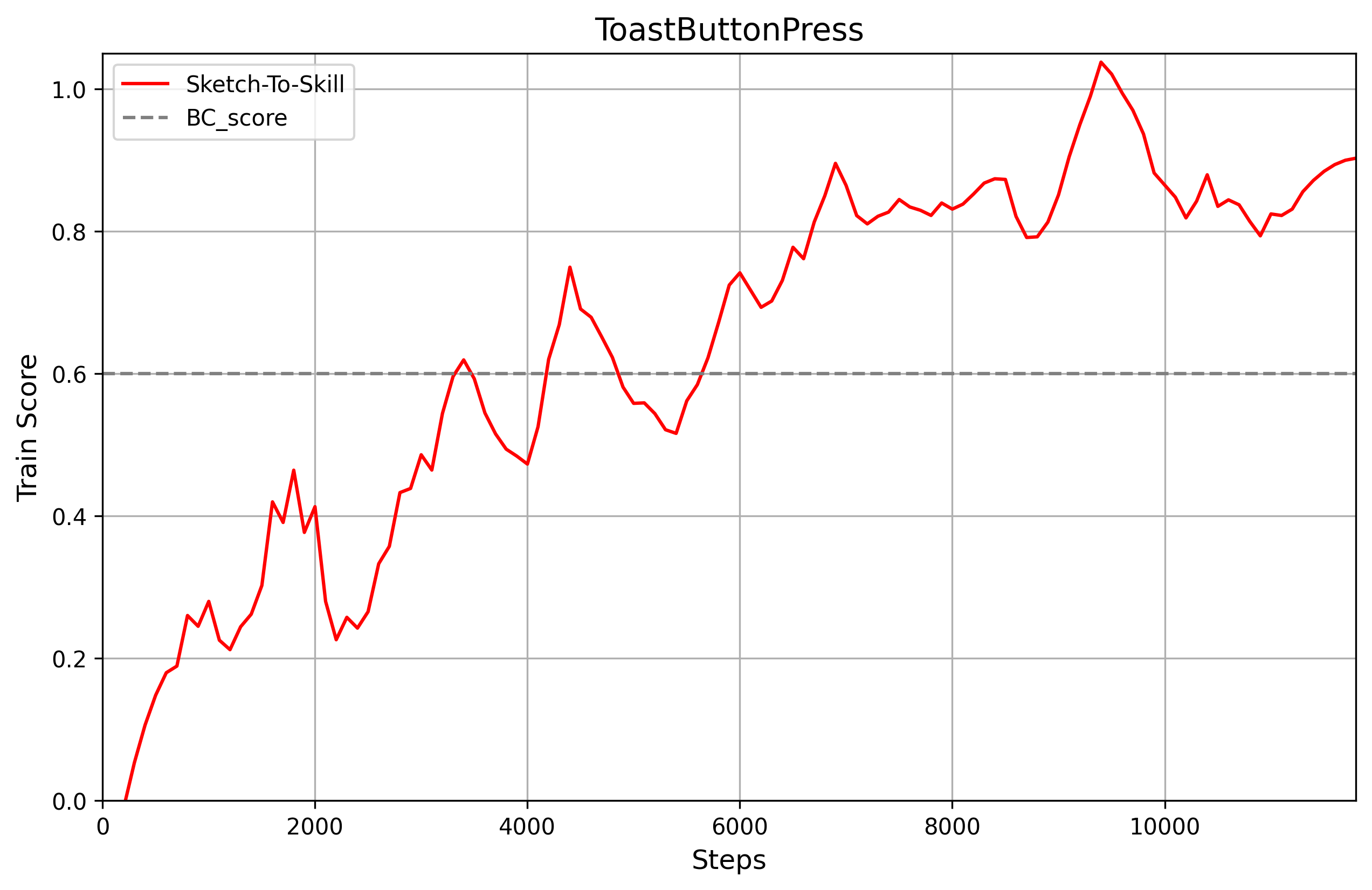}
    \caption{ToastButtonPress.}
    \label{fig:toastpress_setup}
\end{subfigure}

\begin{subfigure}[c]{\columnwidth}
    \centering
    \includegraphics[width=0.8\linewidth]{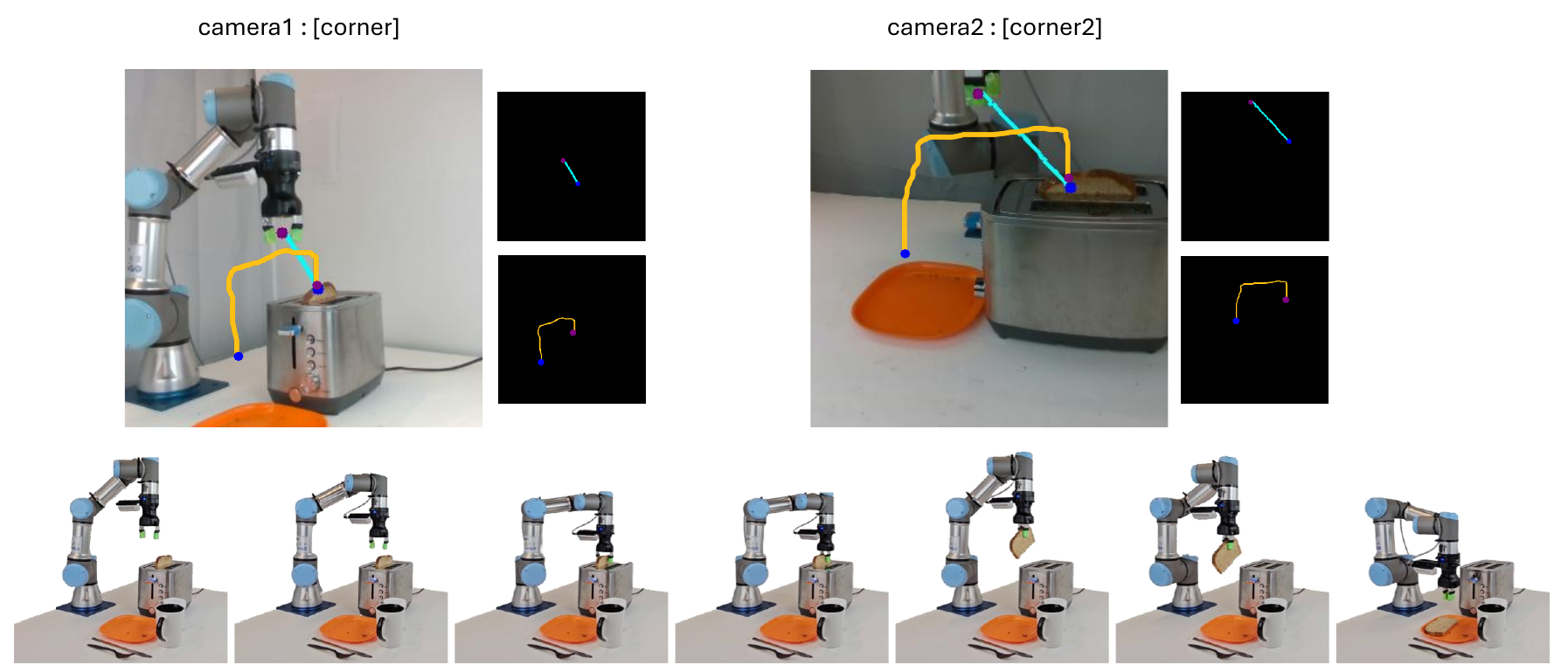}
    % \caption{ToastPickPlace.}
    \label{fig:toastpickplace}
\end{subfigure}
\vspace{1em} % Optional spacing
\begin{subfigure}[c]{\columnwidth}
    \centering
    \includegraphics[width=0.6\linewidth]{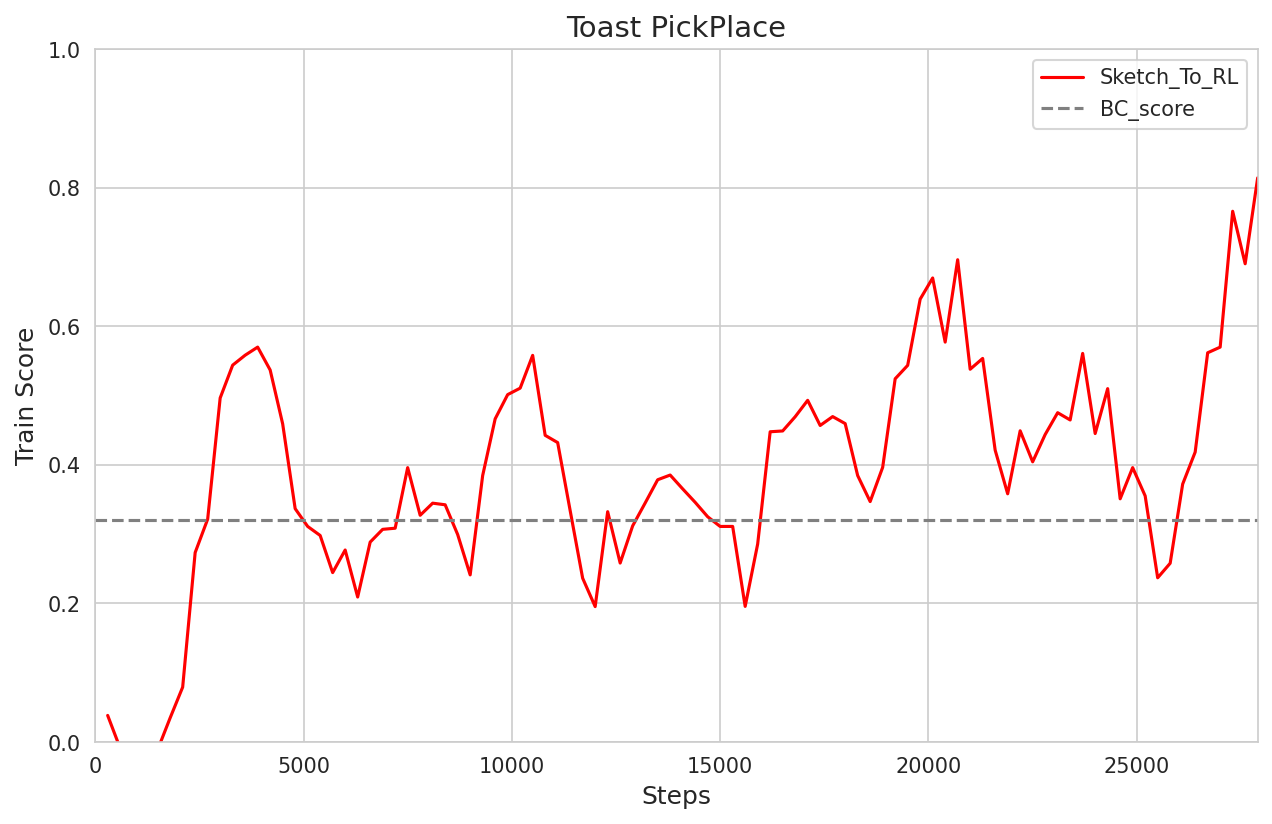}
    \caption{ToastPickPlace.}
    \label{fig:toastpickplace_setup}
\end{subfigure}

\caption{Training curves and hardware setups for real-world robotic tasks: ButtonPress, ToastButtonPress, and ToastPickPlace.}
\label{fig:all_tasks}
\end{figure}

\textbf{Experimental Setup.}
We set up 3 real-world tasks namely \textit{Buttonpress, ToastPress} and \textit{ToastPickPlace} as shown in Figure \ref{fig:all_tasks}. We use a UR3e robot equipped with a Robotiq hand-e gripper and a realsense camera mounted on the wrist. We also use two additional environmental cameras to capture frames for humans to draw sketches on (Figure \ref{fig:exp}). The details of the task, success detection, and reset mechanism are in the Appendix. 
% \vspace{-40pt}

% \PRT{move all specific setup details to the appendix. Just show a rollout and training curve for our algo, not for IBRL, add stats for BC in the caption. might need to move fig 11 to appendix as well}

% \PRT{Amisha/Zahir: photo of the setup, qualitative images of the sketches, snapshots of the policies}

\textbf{Performance.}
The evaluation of our approach across multiple tasks highlights the effectiveness of sketch-generated demonstrations in training policies. For the \textbf{ButtonPress} task, the BC policy achieved a notable success rate of 80\% in a randomized environment, demonstrating the robustness of the approach. Similarly, the Sketch-to-Skill policy, without a discriminator, reached a high training success rate of 80\% within just 30,000 samples. In the \textbf{ToastPress} task, the BC policy achieved a preliminary success rate of 60\% using sketch-generated demonstrations, while the Sketch-to-Skill policy significantly improved performance, achieving approximately 90\% success within 10,000 interactions. Lastly, in the \textbf{ToastPickPlace} task, the BC policy showed an initial success rate of 36\% under dynamic conditions. However, the Sketch-to-Skill policy demonstrated considerable improvement, achieving approximately 80\% success within 30k interactions. 

% The evaluation success rate of the BC policy of ButtonPress task trained on sketch-generated demonstrations is notably high at 0.8 within the randomized environment where we executed the policy. Consequently, our sketch-to-skill policy without a discriminator, quickly demonstrated strong performance, achieving a training success rate of 0.8 within just 30K samples.

\section{Conclusions and Future Work}
We present \OUR{} that uses 2D sketches to improve the efficiency of learning a manipulation skill. While prior work has demonstrated the utility of sketches in imitation learning (IL), we are the first to integrate them effectively within a reinforcement learning (RL) framework. The key ideas involve training a 2D sketch-to-3D trajectory generator, whose output bootstraps RL policy learning and serves as an exploration guidance signal, leading to improved efficiency.

An exciting direction for future work is to explore automated sketch generation from sources beyond human input. For example, \citet{gu2023rt} demonstrated how Vision-Language Models (VLMs) can generate sketches from natural language task descriptions. Integrating such models within our framework could further enhance accessibility and scalability.

\section{Limitations}

While Sketch-to-Skill provides a scalable and accessible approach to bootstrapping robot learning, it is more suitable for shorter horizon tasks where the trajectories to be followed by the robot are intuitive. Our current framework demonstrates success in structured tasks, but extending it to more intricate manipulations warrants further exploration.

Towards addressing this, we have incorporated multi-step task decomposition and RL-guided refinement, improving adaptability in complex scenarios. Enhancing sketches with visual markers and temporal encoding could further refine trajectory representation, particularly for handling overlapping or ambiguous paths.

% Additionally, while sketches offer a practical alternative to teleoperation-based demonstrations, they are best suited as a complementary tool. Future work could explore hybrid approaches that integrate sketch-based guidance with sparse teleoperation data to improve scalability in real-world applications.

% While our Sketch-to-Skill framework offers a scalable and accessible approach to bootstrapping robot learning, it has certain limitations. One key concern is its applicability to complex, long-horizon manipulation tasks requiring fine-grained control, such as dynamic object manipulation and precise end-effector orientation adjustments. Current experiments demonstrate success in structured environments, but extending the approach to more intricate tasks remains challenging.

% To address these limitations, we have explored multi-step task decomposition and reinforcement learning (RL)-guided refinement, which allow for greater adaptability in complex tasks. Additionally, incorporating visual markers and temporal encoding in sketches could improve trajectory representation, enabling better handling of overlapping or partially obscured paths. Experimental validation with color gradients for trajectory segmentation has shown promise in mitigating trajectory ambiguity. Future work can explore integrating adaptive sketch augmentation techniques and expanding the framework to dynamic environments with moving objects.

Furthermore, while sketches provide an accessible alternative to teleoperation-based demonstrations, they do not fully replace high-precision methods. Instead, they serve as an effective complement, especially in resource-constrained settings. Expanding the scope of sketch-based learning to incorporate hybrid approaches—such as combining sketch guidance with sparse teleoperation data—could enhance scalability to more sophisticated robotic applications.

% \section{Reproducibility}
% Anonymized code and demo datasets are be available on \href{https://gsalerts-cyber.github.io/sketch-to-skill/}{our webpage}. We use the standard MetaWorld benchmark to allow for easy comparison with other algorithms and to facilitate the reproducing of our results. All details about the hyperparameters, environment specifications, and real-world experiment setup are provided in the appendix.
%% Use plainnat to work nicely with natbib. 

\bibliographystyle{plainnat}
\bibliography{references}

\clearpage

\section{Appendix}
\subsection{De Boor's Algorithm Details}
\label{appen:deboor}

The uniform knot vector for a B-spline of degree $p$ with $n+1$ control points is defined as:
\begin{equation}
    \vu = [\underbrace{0, ..., 0}_{p+1}, \frac{1}{n-p+1}, \frac{2}{n-p+1}, ..., \frac{n-p}{n-p+1}, \underbrace{1, ..., 1}_{p+1}]
\end{equation}
The B-spline basis functions $\mW_{i,p}(t)$ are defined recursively using the Cox-de Boor recursion formula:
\begin{align}
\mW_{i,0}(t) &= \begin{cases} 1 & \text{if } u_i \leq t < u_{i+1} \\ 0 & \text{otherwise} \end{cases} \\
\mW_{i,p}(t) &= \frac{t - u_i}{u_{i+p} - u_i} \mW_{i,p-1}(t) + \frac{u_{i+p+1} - t}{u_{i+p+1} - u_{i+1}} \mW_{i+1,p-1}(t)    
\end{align}
where $u_i$ are the knot values from the knot vector $\vu$.

The B-spline parametrization matrix $N$ for $m$ evaluation points is an $m \times (n+1)$ matrix:

\begin{equation}
\mW = \begin{bmatrix} 
\mW_{0,p}(t_1) & \mW_{1,p}(t_1) & \cdots & \mW_{n,p}(t_1) \\
\mW_{0,p}(t_2) & \mW_{1,p}(t_2) & \cdots & \mW_{n,p}(t_2) \\
\vdots & \vdots & \ddots & \vdots \\
\mW_{0,p}(t_m) & \mW_{1,p}(t_m) & \cdots & \mW_{n,p}(t_m)
\end{bmatrix}
\end{equation}

where $t_j$ ($j = 1, ..., m$) are evenly spaced parameters in the interval $[0, 1]$.

\subsection{Sketch-to-3D Trajectory Generator Architecture}
\label{appen:traj_app}
\subsubsection*{Overview of the Model}

The proposed model converts 2D image sketches into 3D motion trajectories using a \textit{Variational Autoencoder (VAE)} combined with a \textit{Multi-Layer Perceptron (MLP)}. The VAE encoder processes $64 \times 64$ pixel 2D sketches (3 channels) into a latent vector ($d_{\vv} = 32$), while the decoder reconstructs the sketches to retain essential features for trajectory generation. The latent space outputs the mean $(\mu)$ and variance $(\sigma^2)$, sampled using the reparameterization trick.

The MLP takes the latent vectors from two sketches, concatenates them, and generates 3D control points for B-spline trajectory interpolation. The MLP takes an input of size ($d_{\vv} \times 2$), processes it through hidden layers [1024, 512, 256], and outputs $n_{cp} \times 3$. The generated 3D control points are then used for B-spline interpolation to produce smooth trajectories.

\subsubsection*{Initialization, Regularization, and Hyperparameters}

We initialize all network parameters using \textit{Xavier initialization}. Regularization is done with \textit{Kullback-Leibler Divergence (KLD)}, using a loss function that combines \textit{Sketch Reconstruction Loss}, \textit{KLD Loss}, and \textit{Trajectory Loss}. The \textit{Sketch Reconstruction Loss} is the MSE loss on sketch images, the \textit{Trajectory Loss} computes the MSE between predicted and ground truth trajectories, where the ground truth trajectory is first densely fitted and resampled to ensure uniform point spacing, and the \textit{KLD Loss} applies Kullback-Leibler Divergence for regularization.
Key hyperparameters include: \begin{itemize} \item \textbf{Image size:} $64 \times 64$ pixels \item \textbf{Latent dimension:} 32 \item \textbf{Number of control points:} 20 \item \textbf{B-spline degree:} 3 \item \textbf{MLP hidden layers:} [1024, 512, 256] \item \textbf{Learning rate:} $1 \times 10^{-3}$ (Adam optimizer) \item \textbf{Batch size:} 128 \item \textbf{KLD weight:} 0.0001 (with optional annealing) \item \textbf{Training epochs:} 200 \end{itemize}

\begin{figure}[h]
% \vspace{-0.5cm}
    \centering
    \includegraphics[width=0.95\linewidth]{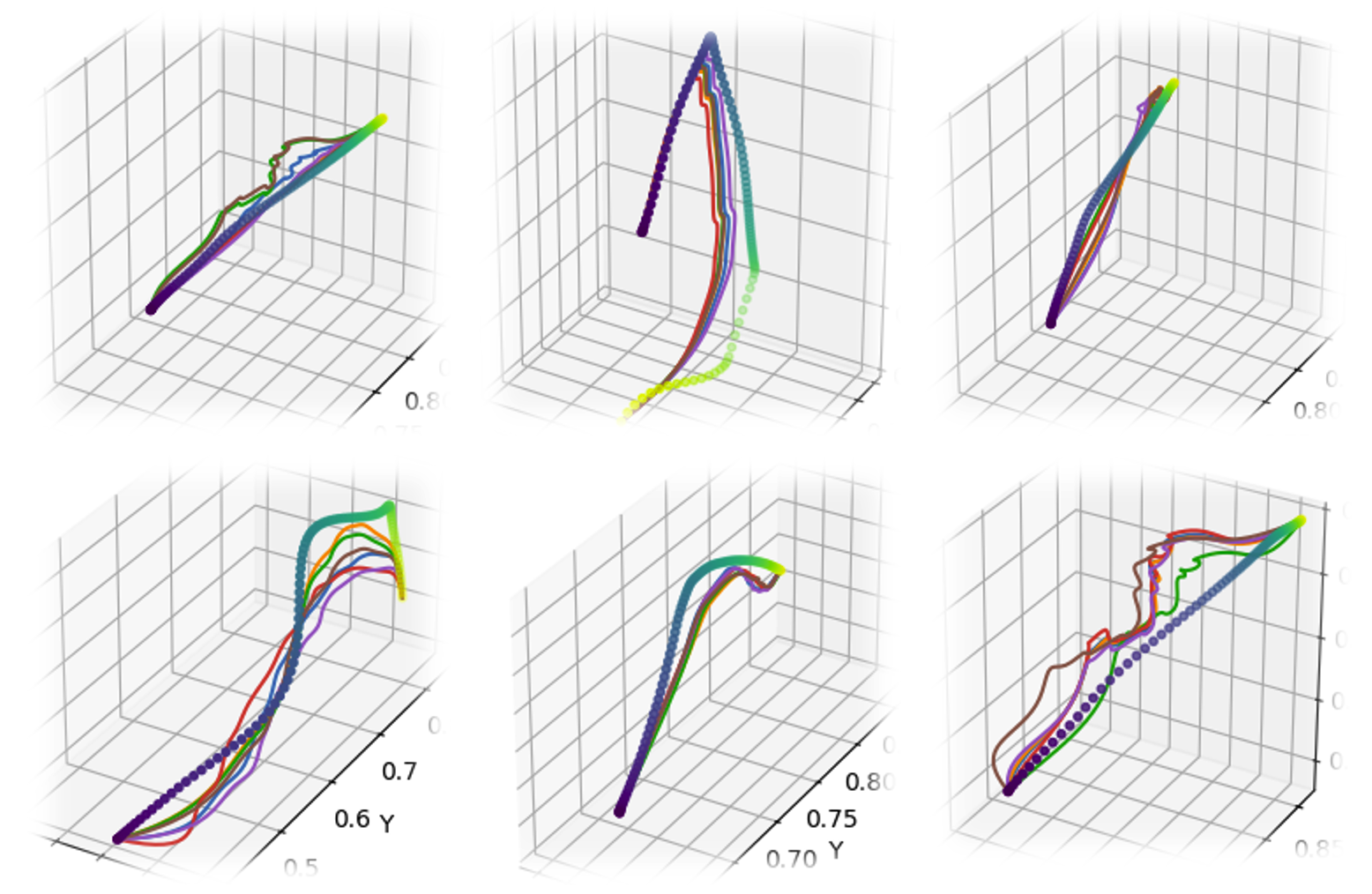}
    \caption{Diversity in generated 3D trajectories. Each subplot shows multiple generated trajectories (colored lines) for a single input, demonstrating variability. Scattered points represent the ground truth trajectory.}
    \label{fig:diversity}
% \vspace{-0.5cm}
\end{figure}
\begin{figure*}[h]
    \centering
    \includegraphics[width=1\linewidth]{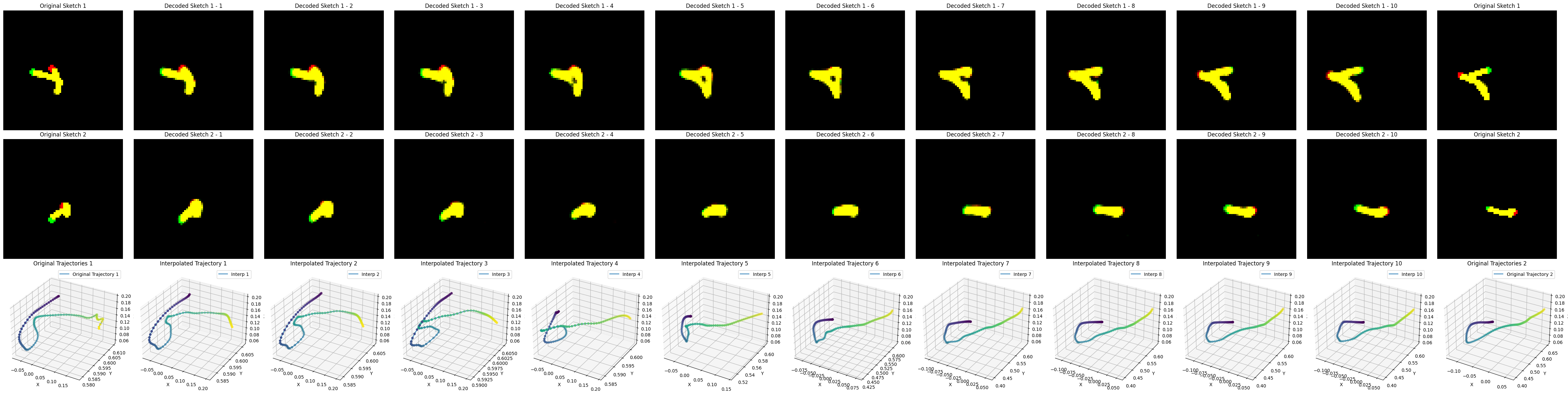}
    \caption{Latent space interpolation results showing smooth transitions between original samples (leftmost and rightmost) and reconstructed samples (middle) from two viewpoints, demonstrating the model's ability to generate coherent 3D trajectories from 2D sketch pairs. }
    % \yuph{several other examples in the "interpolated" folder, change if you think others are bette.}}
    \label{fig:interpolated}
\end{figure*}

To enhance robustness and generalization, our training process employs two concurrent data augmentation strategies. The first applies diverse image augmentations (rotations, scaling, affine transformations, noise) to input sketches, used exclusively for updating the VAE to learn robust sketch representations. The second strategy targets potential mismatches in hand-drawn sketches by subtly modifying both original sketches and their 3D trajectories. This involves adding noise and minor elastic deformations to sketches, and noise with refitting to trajectories. These augmented pairs update the entire model, preparing it for hand-drawn input variability while maintaining sketch-trajectory consistency. This augmentation approach enhances the model's ability to handle diverse, imperfect sketches while ensuring accurate 3D trajectory generation in real-world scenarios. To train the sketch generation network we collect $22000$ samples for simulation tasks and $85$ for the real world. During the training for the real-world task, we use the data collected from simulation to train the VAE part as well. We provide additional visualizations to demonstrate the learned model's generalizability in Figure \ref{fig:diversity} and \ref{fig:interpolated}.
% \yuph{how many samples are used for training in simulation and in real world, losses}

\subsection{Implementation Details and Hyperparameters of \OUR{} Policy and baselines}

This section outlines the implementation details of \OUR{} and the baselines. The behavior cloning (BC) policies utilize a ResNet-18 encoder, where the output is flattened and processed by MLPs to produce final 4D actions. We replace the BatchNorm layers in ResNet with GroupNorm, matching the number of groups to the input channels. To prevent overfitting, we employ random-shift data augmentation.

In the Meta-World environment, we utilize a corner2-image camera setup. We use wrist cameras to enhance generalization and sample efficiency in real-world experiments, specifically using them for the ButtonPress task.

\begin{table}[ht]
    \centering
    \caption{Hyperparameters for RL in \OUR{}.}
    \begin{tabular}{|l|c|c|c|c|}
        \hline
        \textbf{Parameter} & \textbf{Meta-World}   & \textbf{Real-World} \\
        \hline
        Optimizer & Adam & Adam  \\ 
        Learning Rate & 1e-4 & 1e-4  \\
        Batch Size & 256 & 256 \\
        Discount ($\gamma$) & 0.99 & 0.99  \\
        Exploration Std. ($\sigma$) & 0.1 & 0.1  \\
        Noise Clip ($c$) & 0.3 & 0.3  \\
        EMA Update Factor ($\rho$) & 0.99 & 0.99 \\
        Update Frequency ($U$) & 2 & 2  \\
        Actor Dropout & 0.5 & 0.5  \\
        Q-Ensemble Size ($E$) & 2 & N/A  \\
        Num Critic Update ($G$) & 1 &  N/A  \\
        Image Size & 96×96 & 96×96  \\
        Use Proprio & No &  N/A  \\
        Proprio Stack & N/A &  N/A  \\
        State Stack & N/A  & N/A  \\
        Action Repeat & N/A  & N/A \\
        \hline
    \end{tabular}
    \label{tab:hyperparameters}
\end{table}

\subsection{Additional Details of Real-world Experiments}

In this section, we present insights into the real-world experiments conducted using \OUR{}. We utilized a UR3e robot equipped with a Robot Hand gripper, operating in an action space with 4 dimensions: 3 for end-effector position deltas under a Cartesian impedance controller and 1 for the absolute gripper position, with policies functioning at 7.5 Hz.

To train the Sketch-To-3D Trajectory generator, we collected approximately 85 teleoperated trajectories. RGB images were captured using two orthogonally positioned RealSense cameras to enhance trajectory insight. A green marker was placed on the gripper tip to facilitate sketch generation on the frames, enabling the model to learn 2D sketch projections onto 3D trajectories. All methods maintained the same hyperparameters and network architectures as those used in the Meta-World tasks.

\subsubsection{Common Implementation Details}

\begin{itemize}
    \item \textbf{Reset Protocol: }
For all tasks, we manually reset the environment between episodes by returning objects to their initial states and randomizing their positions. The robot is initially set to a specific joint configuration known as the home position. Whenever the agent receives a reward or an episode ends, it resets back to the home configuration and the training continues. The object randomization typically begins with placement at the center, remaining unchanged until the agent receives its first reward. After that, the object is gradually moved towards the boundary, circled around the workspace, returned to the center, and the process is repeated until the training ends.
\item \textbf{Safety Boundaries: }
We have restricted the movement of the robot to the x, y, and z directions of the end-effector. Each step that the robot takes is limited to a specific value. If the action taken by the agent exceeds this limit, the robot will not move and will remain in its current position. To avoid collision with the workspace surface, we have set a minimum limit for the z direction of the robot. Even if the agent attempts to move downward beyond this threshold, the robot will remain at the specified z position.
\item \textbf{Reward Structure: }
We used a manual method to reward the agent if it successfully completes the task. The agent receives a reward of 1 for successfully completing the task and a reward of 0 in all other cases. Each episode has a set number of timesteps, and if the agent doesn't succeed within that limit, it resets and starts over. The length of each episode may be shorter than the given limit depending on how quickly the agent completes the task.
\end{itemize}

\subsubsection{Task-Specific Details}

\begin{itemize}
    \item \textbf{Button Press Task: }
After training the Sketch-To-3D Trajectory generator, we created 30 sketches based on RGB frames from the two cameras. We then collected 30 sketch-generated demonstrations, $\xi_D$ using openloop servoing on 3D trajectory $\xi_g$ produced by the generator $\mT$. We then train Behavior Cloning (BC) policy using the sketch-generated demonstrations, $\xi_D$, achieving a score of 0.8 and thus leading to good performance in the \OUR{} policy. The button position was randomized within a 20cm by 20cm to 25cm trapezoidal area visible from the wrist camera.

\item \textbf{Toaster Press Task: }
The Toaster Press experiment was designed to test the robustness of our approach in cluttered environments. We created 10 sketches based on the camera feeds and collected 10 sketch-generated demonstrations using open-loop servoing. Each episode featured a cluttered environment around the toaster, with objects commonly found in household settings to simulate realistic conditions. The initial position of the gripper was randomized in every episode to test the adaptability of the learned policy. Training a Behavior Cloning (BC) policy using the sketch-generated demonstrations achieved a preliminary success rate of 60\%, and \OUR{} training over 12k interactions achieved $\sim$90\% success rate within 10k interactions.

\item \textbf{Bread Pick and Place Task}
This task involved picking a piece of bread from a toaster and placing it on a nearby plate, requiring precise manipulation and handling. Sketches were specifically collected in randomly cluttered environments to reflect typical variability in real-world scenarios. Both the environment clutter and the initial gripper positions were randomized in each episode, presenting a different challenge each time to test the robustness of the policy. The BC policy achieved a success rate of 36\% and \OUR{} trained over 30K interactions achieved $\sim$80\% success within 30K interactions.

\end{itemize}

These experiments underscore our method's capability to handle real-world variability and complex task execution, supporting its potential utility beyond controlled experimental setups. The detailed results from these tasks, illustrated in Figures 16 and 17, highlight the practical applications and adaptability of our \OUR{} framework in dynamic, cluttered settings.

% The task is illustrated in Figure \ref{fig:hardware_scores} and briefly described below:

\begin{figure*}
    \centering
    \includegraphics[width=0.8\linewidth]{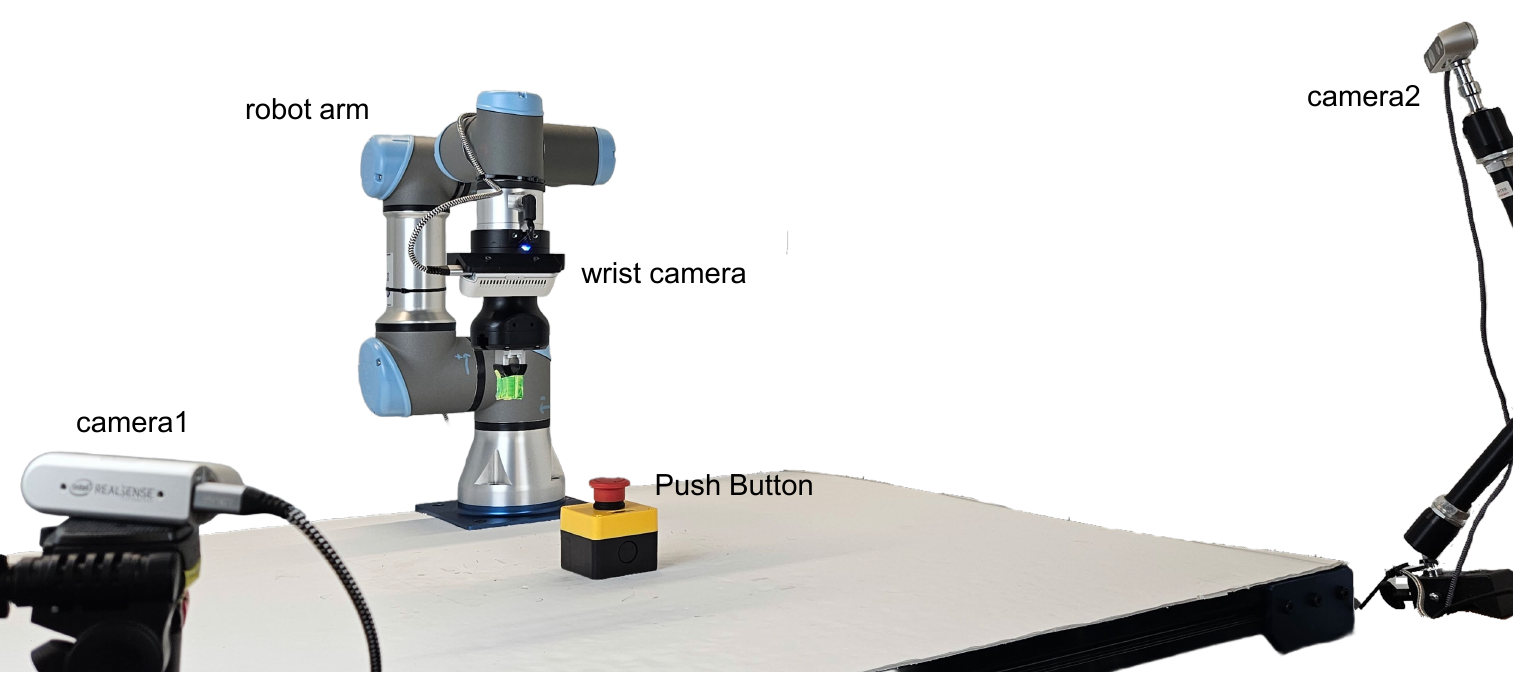}
    \caption{Complete setup for the ButtonPress task in a real-world experiment. The configuration includes a UR3e robot arm equipped with a Robot Hand gripper, and a RealSense D435i camera mounted on the wrist. Two additional RealSense cameras are positioned orthogonally to capture the trajectory from two different viewpoints.}
    % \yuph{several other examples in the "interpolated" folder, change if you think others are bette.}}
    \label{fig:exp}
\end{figure*}

\subsection{Additional Details of Robomimic Experiments}
\subsubsection{Task Complexity} The PickPlaceCan task in RoboMimic is an demanding two-stage challenge. It requires the robot to accurately locate and reach a can, then pick it up and place it into a designated bin. This task not only tests the robot's ability to handle objects with precision but also demands correct orientation of the gripper throughout the process. RoboMimic, a well-established benchmark, provides high-quality demonstrations collected via human teleoperation, which are instrumental for training successful policies.

\subsubsection{Implementation and Strategy} In this setup, the sketch-based demonstrations provide initial positional cues for the robot. Our framework utilizes reinforcement learning (RL) to refine these initial cues, dynamically adjusting the robot’s approach to manage both the orientation and timing necessary for successful task execution. This method proves particularly effective as it does not rely on fully detailed trajectory information from the start; instead, it uses the sketches to guide the initial exploration phase of RL, significantly simplifying the data collection process.

\subsubsection{Results and Performance Metrics}
\begin{itemize}
    \item  \textbf{Performance:} Our \OUR{} framework, even with limited initial data, performs admirably in this demanding scenario. The results are on par with those from IBRL methods ~98\%, which benefit from complete and detailed human demonstrations, including explicit orientation details (as shown in Figure 15).
    \item \textbf{Comparison to IBRL:} The comparative success illustrates the robustness and effectiveness of our method in managing the task's orientation and other complexities without fully specified trajectory inputs. This is particularly notable given that RoboMimic's Pick and Place Can task offers a significantly higher level of difficulty than similar tasks in MetaWorld.
\end{itemize}

\subsection{Additional Experiments in Metaworld Assembly}

\begin{figure*}[h!]
\begin{subfigure}[c]{0.99\columnwidth}
\centering
    \includegraphics[width=\textwidth]{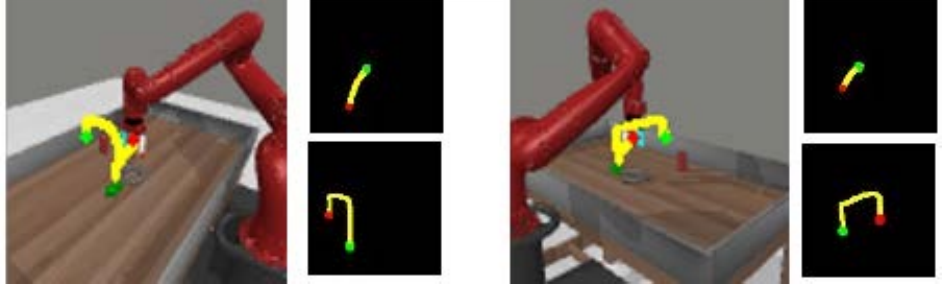}
\end{subfigure}
\centering
\begin{subfigure}[c]{0.79\columnwidth}
\centering
    \includegraphics[width=1\linewidth]{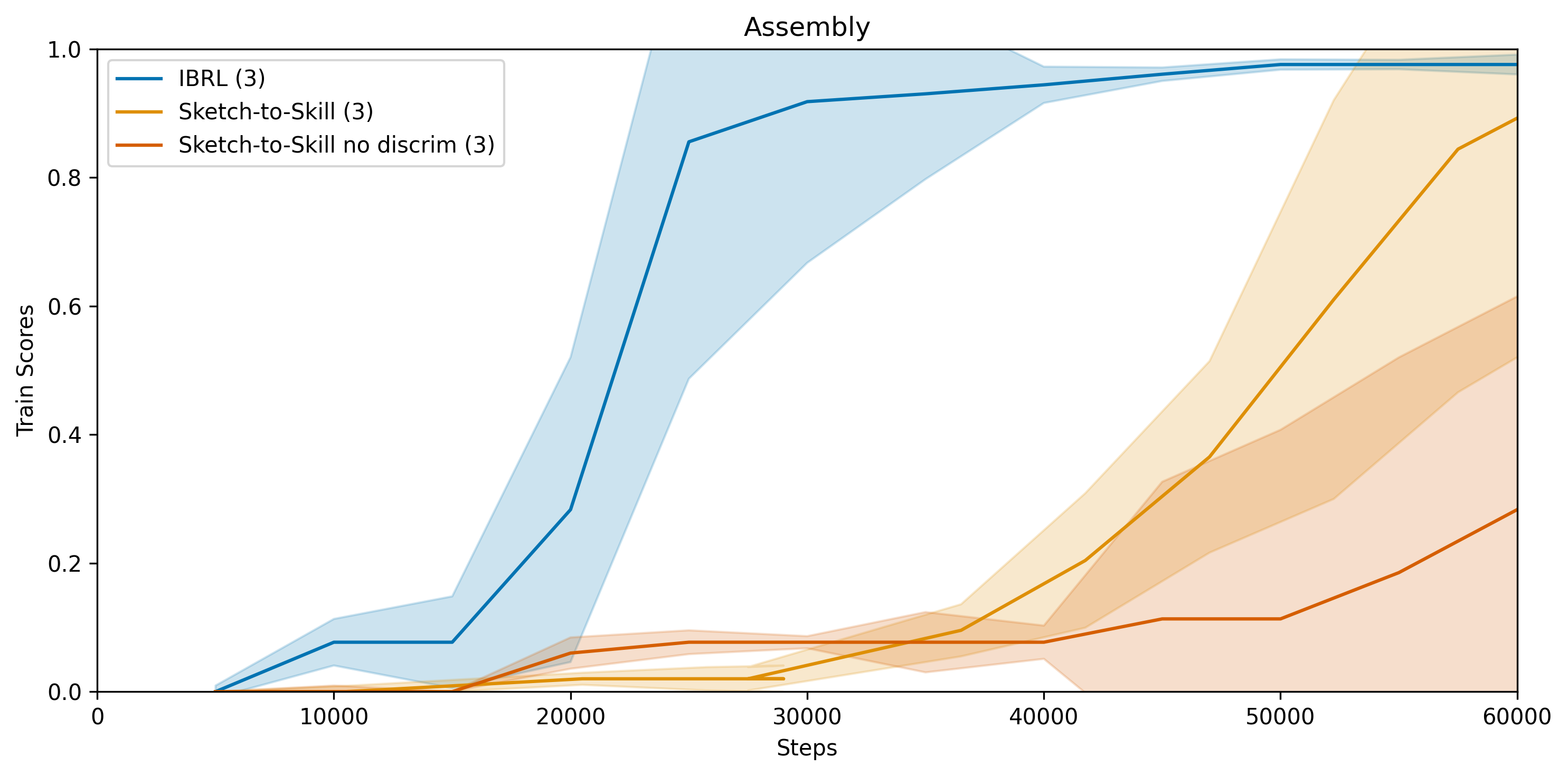}
\end{subfigure}
% \begin{subfigure}[c]{0.49\columnwidth}
% \centering
%     \includegraphics[width=0.98\textwidth]{figures/output.png}
% \end{subfigure}
\caption{Left: sketches for the MetaWorld Assembly task. Right: training scores (success rate).}
\label{fig:assembly}
\end{figure*}
% \begin{figure}[h!]
%     \centering
%     \includegraphics[width=0.5\linewidth]{figures/Assembly.png}
%     \caption{Training scores (success rate) for the Metaworld Assembly task.}
%     \label{fig:assembly}
% \end{figure}

% MetaWorld Assembly Task Description

\subsubsection{Task Complexity} The Assembly task in MetaWorld is a ``hard'' task~\cite{seo2023masked} that requires the robot to execute two-stage manipulations. The task involves precise movements to pick up a peg, navigate it to a specific location, and insert it into a hole within a larger assembly fixture. This task tests the robot's precision, spatial awareness, and ability to handle complex sequences.
\subsubsection{Results and Performance Metrics}
\begin{itemize}
    \item  \OUR{}: Demonstrated a high success rate of 93\%, effectively using sketches for coarse guidance to navigate and complete complex task sequences, even without actual teleoperated demonstrations.
    \item \OUR{} without Discriminator: Initially struggled with a success rate of 20\%, but extended interaction up to 100K steps improved performance dramatically, reaching a near-perfect success rate of 98\%. This underscores the potential for learning even without discriminator guidance, given sufficient training time.
    \item IBRL: Achieved the highest success rates of approximately 100\%, benefiting from high-quality teleoperated demonstrations that include precise details on orientation and positioning.
    \item Standard BC: Achieves a success rate of around 60\%, showing limitations in environments where adaptive behaviors and fine-tuning through reinforcement learning are necessary.
    
\end{itemize}

These results highlight the effectiveness of the \OUR{} approach in handling complex, multi-stage tasks through initial coarse guidance, with the potential for significant improvement over time. They also illustrate the advantage of incorporating discriminator feedback to accelerate learning and enhance performance.
% MetaWorld Assembly Task Description
% \color{red}

\begin{figure*}
\centering
    \includegraphics[width=0.9\linewidth]{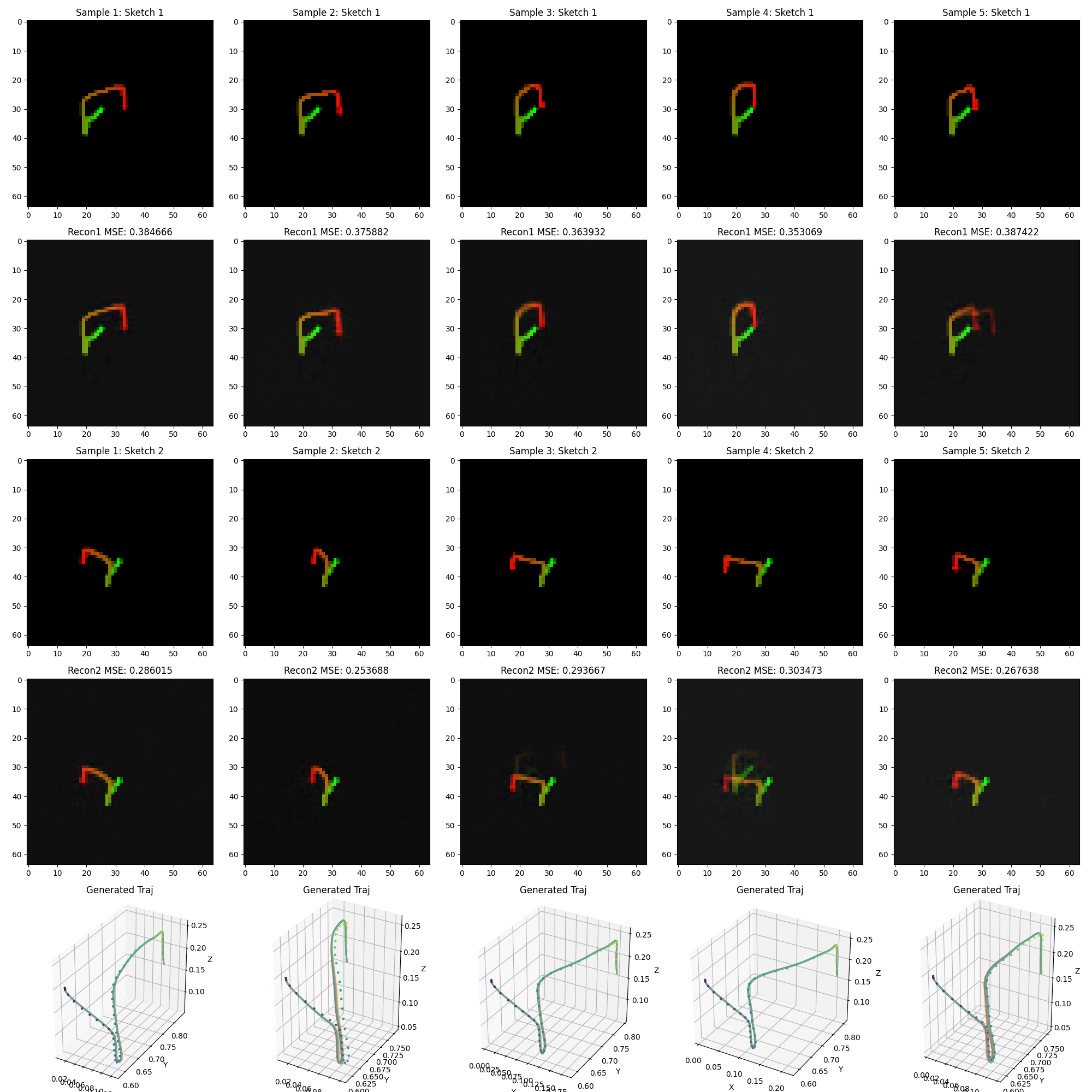}
\caption{Trajectories with time-based color gradient for Assembly Metaworld simulation task.}
\label{fig:generator}
\end{figure*}

\subsection{Using Color Gradients for Overlapping Trajectories}
We can also incorporate time-parameterization of the trajectory in the sketches using a color gradient, instead of a binary sketch image. This notion of color gradient in sketches was introduced by RT-Trajectory~\cite{gu2023rt}. An example is shown in Figure~\ref{fig:generator}. Here, the color of the sketch changes from green (start) to red (end). This is particularly helpful when the sketch crosses each other or overlaps. We conduct additional experiments with the generator to evaluate the effect of incorporating gradients in the Sketch-To-3D trajectory generator. 

Specifically, we use the MetaWorld Assembly task (Figure~\ref{fig:assembly}) where the sketch overlaps in the middle as the end-effector picks up the tool and carries it to the goal position. We trained two generators without and with color gradients. The performance of these two generators are reported in Table~\ref{tab:overlap}. We observe that incorporating the gradient in such a case results in lower reconstruction and trajectory losses. With lower trajectory losses we can handle more complex trajectories that overlap with color gradients. Note that incorporating gradients also does not require any change to the downstream architecture, and only requires minimal changes to the generator architecture.

\begin{table}[ht]
\centering
\caption{Performance Metrics for Overlapping and Non-Overlapping Trajectories}
\label{tab:overlap}
\begin{tabular}{
  @{}l 
  % >{\raggedright\arraybackslash}p{3.0cm} 
  % >{\raggedright\arraybackslash}p{3.0cm} 
  >{\raggedright\arraybackslash}p{3.0cm} 
  >{\raggedright\arraybackslash}p{3.0cm}@{}
}
\toprule
Metric   &  with color gradient &  without color gradient \\ \midrule
Training Loss           & \textbf{0.2438}  & 0.3107\\
Validation Loss         & \textbf{0.3120}  & 0.3585\\
Reconstruction Loss     & \textbf{0.0001} & 0.0003 \\
KLD Loss                & \textbf{77.167} & 76.4683 \\
Parameter MSE Loss      & \textbf{0.1028} & 0.1135 \\
Trajectory loss         & \textbf{0.1530} & 0.2379 \\ \bottomrule
\end{tabular}
\end{table}

% \begin{figure}[h!]
% \centering
%     \includegraphics[width=0.5\textwidth]{figures/aug_sketch.png}
% \caption{Augmented Trajectories}
% \label{fig:generator}

% \end{figure}
% \begin{figure}[h!]
% \centering
%     \includegraphics[width=0.98\textwidth]{figures/aug.png}
% \caption{Augmented Sketches for training generator model}
% \label{fig:generator}
% \end{figure}

% overlap, gradient
% no, no - 0.31, 0.35
% no, yes - 

% yes, no - 0.31, 0.378
% yes, yes - 0.24, 0.3120

% vae, aug
% no, no - 0.628, 1.2
% yes, no - 0.60, 0.9028

% no, yes - 2.28, 2.69
% yes, yes - 4.28, 5.9

\end{document}